\documentclass[10pt,twocolumn,letterpaper]{article}

\usepackage[pagenumbers]{cvpr} 

\definecolor{cvprblue}{rgb}{0.21,0.49,0.74}
\usepackage[pagebackref,breaklinks,colorlinks,allcolors=cvprblue]{hyperref}

\usepackage{graphicx}
\usepackage{subcaption}  
\usepackage{tabularx}    
\usepackage{booktabs}
\usepackage{tcolorbox}
\usepackage{algorithm}
\usepackage{algpseudocode} 

\usepackage{algpseudocode}
\usepackage[rgb]{xcolor}
\usepackage{multirow}
\usepackage{wrapfig} 
\usepackage{caption}
\usepackage{makecell} 
\usepackage{xspace}
\usepackage{float}
\usepackage{placeins}
\usepackage{afterpage}

\newcommand{\alglongname}{InstructMix2Mix\xspace}
\newcommand{\algname}{I-Mix2Mix\xspace}
\newcommand{\NA}{\makebox[1.2cm][c]{--}}

\title{\alglongname: Consistent Sparse-View Editing Through Multi-View Model Personalization}

\author{
Daniel Gilo$^{1}$ \quad Or Litany$^{1,2}$\\[0.3em]
$^{1}$Technion -- Israel Institute of Technology\\
$^{2}$NVIDIA\\[0.3em]
{\tt\small danielgilo@cs.technion.ac.il \quad or.litany@gmail.com}
}

\algrenewcommand\algorithmiccomment[1]{\hfill\(\triangleright\) #1}

\begin{document}
\maketitle

\begin{abstract}
We address the task of multi-view image editing from sparse input views, where the inputs can be seen as a mix of images capturing the scene from different viewpoints. The goal is to modify the scene according to a textual instruction while preserving consistency across all views.
Existing methods, based on per-scene neural fields or temporal attention mechanisms, struggle in this setting, often producing artifacts and incoherent edits. We propose \alglongname (\algname), a framework that distills the editing capabilities of a 2D diffusion model into a pretrained multi-view diffusion model, leveraging its data-driven 3D prior for cross-view consistency. A key contribution is replacing the conventional neural field consolidator in Score Distillation Sampling (SDS) with a multi-view diffusion student, which requires novel adaptations: incremental student updates across timesteps, a specialized teacher noise scheduler to prevent degeneration, and an attention modification that enhances cross-view coherence without additional cost. Experiments demonstrate that \algname significantly improves multi-view consistency while maintaining high per-frame edit quality. Additional visualizations and code are available on our \href{https://danielgilo.github.io/instruct-mix2mix/}{project page}.
\end{abstract}

\section{Introduction}
Multi-view image editing seeks to modify a scene captured from multiple viewpoints while preserving consistency across views. Typical edits include texture and color changes, semantic manipulations, or geometric transformations, with applications in product imagery, real-estate and interior design visualization, AR/VR, and cinematic post-production.

However, multi-view editing is a difficult task that traditionally requires skilled artists, making automation highly desirable. In recent years, a variety of methods have been proposed for editing 3D scenes. Due to the difficulty of producing supervision in the form of high-quality paired scenes before and after editing, most of these approaches avoid direct multi-view editing, instead leveraging monocular editors such as InstructPix2Pix \cite{brooks2023instructpix2pix}, either iteratively \cite{haque2023instruct, vachha2024instruct} or through distillation \cite{li2024focaldreamer, kamata2023instruct}. These approaches achieve 3D-consistent edits by operating on dense scene representations like NeRFs \cite{mildenhall2021nerf} or 3D Gaussian Splats \cite{kerbl3Dgaussians}. 
However, in many real-world scenarios, users have access only to a sparse set of views—such as a few casually captured photos or product images—which provide limited scene coverage and challenge existing 3D editing methods to maintain geometric and appearance consistency.

In this work, we tackle the challenging problem of multi-view editing from sparse input images (a \emph{mix} of images). Given a few source views and a textual editing instruction, our aim is to generate edits that faithfully follow the prompt while remaining consistent across all viewpoints.
Following prior work, we leverage powerful 2D editors and lift their capabilities to 3D using Score Distillation Sampling (SDS) \cite{poole2022dreamfusion}. However, we propose a new approach to this paradigm.
We observe that current approaches
face inherent limitations. Neural field representations are trained per scene --  they do not hold 3D prior in their network weights. Instead, they achieve 3D consistency by incorporating a physical prior through the rendering equations. A \emph{dense set} of  input images is required, however, to transform this prior into an effective consolidator. To achieve consistency with \emph{sparse} views, we instead propose to incorporate a consolidator that embeds a strong, data-driven 3D prior directly in its weights: a multi-view synthesis diffusion model. While such models are trained to generate view-consistent scenes (e.g., from text or images), they lack editing capabilities. We bridge this gap by combining the strengths of both paradigms—distilling edits from a 2D editor (the teacher) into the multi-view model (the student). Concretely, we use InstructPix2Pix as the teacher and Stable Virtual Camera (SEVA) \cite{zhou2025stable} as the student. By leveraging a student model with an inherent 3D prior, our method —  \textbf{\algname} — produces robust, geometrically coherent, and visually consistent edits even from extremely sparse inputs.

Replacing the neural field with a multi-view diffusion model within the SDS framework is not straightforward, and requires careful adaptation of several key steps. Instead of rendering from a scene representation, we sample from the diffusion model; to avoid costly full trajectories, we distill incrementally across student timesteps. We also introduce a specialized teacher noise scheduler to prevent collapse to poor local minima and an attention modification that strengthens multi-view consistency without extra cost. Together, these components yield a framework for consistent multi-view editing, producing high-quality results even with very sparse inputs.

\noindent
To summarize, our contributions are:  
\begin{enumerate}
    \item We present \algname, a novel framework for distilling the knowledge of a powerful monocular editor into a pretrained multi-view diffusion model, leveraging its data-driven 3D consistency.
    \item Through careful consideration of the SDS key steps, we introduce novel adaptations to support personalization of our multi-view student. 
    \item We demonstrate that this approach produces high-quality, consistent multi-view edits, effectively extending the SDS framework to scenarios with limited viewpoints.
\end{enumerate}

We evaluate \algname against popular multi-view editing methods, demonstrating significant improvements in cross-view consistency both qualitatively and quantitatively in the sparse-view setting. At the same time, our method maintains competitive per-frame editing performance, highlighting the practical benefits of leveraging a data-driven multi-view prior within the SDS framework.

\section{Related Works}
\label{related_works}

\paragraph{3D editing.}  
Editing 3D scenes or objects typically assumes a pre-optimized model such as a NeRF \cite{mildenhall2021nerf} or 3DGS \cite{kerbl3Dgaussians}. Early works explored \emph{direct NeRF manipulation} via scribbles \cite{liu2021editing}, sketches \cite{mikaeili2023sked}, reference images \cite{bao2023sine}, meshes \cite{yuan2022nerf}, point clouds \cite{chen2023neuraleditor}, and other cues \cite{weder2023removing, mirzaei2023spin, yang2021learning}, while \emph{NeRF stylization} transfers reference appearances to 3D scenes \cite{wang2022clip, wang2023nerf, nguyen2022snerf, huang2022stylizednerf, chiang2022stylizing}.  
\emph{Instruction-based} approaches leverage 2D diffusion editors like InstructPix2Pix \cite{brooks2023instructpix2pix}, applying SDS-like guidance \cite{li2024focaldreamer, sella2023vox, zhuang2023dreameditor, kamata2023instruct, huang2025dacapo} or Iterative Dataset Update \cite{haque2023instruct, wang2024proteusnerf, vachha2024instruct, wang2024gaussianeditor, chen2024gaussianeditor, chen2024generic} for improved consistency. Some methods \cite{gomel2024diffusionbasedattentionwarpingconsistent, dong2023vica, li2025syncnoise} further enforce alignment using depth information from the trained NeRF, while others \cite{chen2023shapeditor, xia2025towards, parelli20253d} operate directly in a learned 3D latent space. While effective for 3D editing, the reliance of these approaches on dense input views or an initial 3D representation makes them less suitable in sparse-view scenarios.

\vspace{-1em}
\paragraph{Sparse multi-view editing.}
In the absence of a full 3D representation, several works have explored editing a set of input images directly. A prominent direction adapts pre-trained diffusion-based monocular editors by modifying self-attention layers: as first shown in \citet{wu2023tune}, extending queries to attend across frames promotes consistency between the outputs. Building on this idea, a number of methods generate edited image sequences \cite{geyer2023tokenflow, shin2024edit, khachatryan2023text2video, ceylan2023pix2video, qi2023fatezero, liu2024video, zhu2025coreeditor}. While effective for temporally smooth sequences with small viewpoint changes, these approaches struggle in the sparse-view setting, where edits must remain consistent under significant viewpoint differences.
DGE \cite{chen2024dge} combines extended attention with 3D Gaussian Splatting (3DGS) lifting: attention-based edits provide rough multi-view consistency, while 3DGS is used to consolidate outputs and resolve residual artifacts. However, in the sparse-view regime, 3DGS tends to overfit the limited input views rather than serve as a true cross-view aggregator \cite{zhang2024cor, zhu2024fsgs}, leading to persistent inconsistencies. As a result, DGE effectively reduces to an extended-attention approach, inheriting the same limitations as prior video editing methods.
Most recently, \citet{bar2025editp23} proposed a feed-forward approach that propagates a user-provided 2D edit to multiple views, but their method remains limited to object-level edits.
Contemporaneously with our work, \citet{zhao2025tinker} fine-tune FLUX Kontext \cite{labs2025flux} to enable consistent edits across image pairs, while \cite{chi2025disco3d,tao2025c3editor} distill 3D consistency priors into a 2D editor.

\begin{figure*}[t]
    \centering
    \includegraphics[width=1\linewidth]{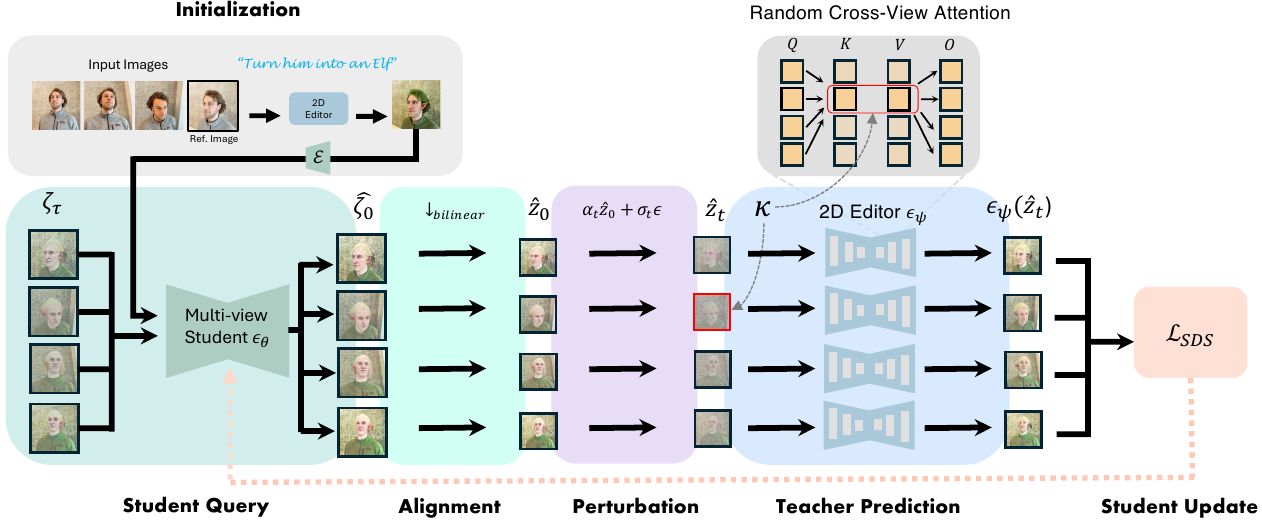}
    \caption{
    \textbf{\algname\ overview}. 
Given a set of input images, a randomly chosen reference image is edited by the frozen teacher and encoded to serve as the personalized multi-view student's input latent (\emph{Initialization}). 
At each distillation iteration, noisy multi-view latents $\zeta_\tau$ are denoised by the student (\emph{Student Query}), aligned to the teacher’s latent space (\emph{Alignment}), and perturbed with our forward schedule (\emph{Perturbation}). 
The teacher predicts edits with Random Cross-View Attention (\emph{Teacher Prediction}), where all frames attend to the $\kappa$'s frame, and the resulting supervision is distilled back into the student (\emph{Student Update}). After distillation, the student outputs a set of multi-view consistent edited frames.  
} 

    \label{fig:method}
\end{figure*}

\section{Preliminaries}
\label{preliminaries}

\paragraph{Stable Virtual Camera.} SEVA \cite{zhou2025stable} is a diffusion-based Novel View Synthesis (NVS) model that predicts $N$ target images given $M$ input images with their camera poses. Built on Stable Diffusion 2.1 \cite{rombach2022high} with architectural adaptations for NVS and trained on diverse object and scene datasets, it achieves state-of-the-art results, making it an ideal student model with a strong 3D prior.
\vspace{-1em}
\paragraph{Instruct-Pix2Pix.} 
A monocular, instruction-based image editing diffusion model widely used in 3D and multi-view editing, Instruct-Pix2Pix \cite{brooks2023instructpix2pix} is fine-tuned from a pre-trained Stable Diffusion model on a large-scale synthetic editing dataset. Given a source image and a textual instruction, it produces versatile edits by sampling the fine-tuned model. The model employs classifier-free guidance (CFG) \cite{ho2022classifier} with two scales: a \emph{text CFG scale} $s_T$ controlling adherence to the instruction, and an \emph{image CFG scale} $s_I$ controlling fidelity to the source image, jointly balancing edit strength and overall image quality.

\section{Method}
\label{sec:method}

Our goal is sparse multi-view consistent image editing. We build on the SDS framework~\cite{poole2022dreamfusion}, using a pre-trained image editing network as a \emph{teacher} to distill knowledge into a neural scene representation \emph{student}. Unlike typical settings that assume abundant input views, we work with only a few images. To address this challenge, we replace the conventional neural field with a multi-view diffusion model pre-trained for consistent view generation. We personalize this student to the target scene and edit instruction by distilling the teacher’s predictions, enabling faithful and consistent edits from limited inputs.

\subsection{Problem Formulation}

We are given $N$ images $\{I_i\}_{i=1}^N, \; I_i \in \mathbb{R}^{3 \times H \times W}$ of a static 3D scene with camera poses $\{\pi_i\}_{i=1}^N, \; \pi_i \in \mathbb{R}^{4 \times 4}$, and an editing prompt $y \in \mathcal{Y}$. We assume access to a multi-view  diffusion model $\epsilon_\theta$ which we refer to as the \emph{student}, and a monocular instruction-based editing diffusion model (\emph{teacher}) $\epsilon_\psi$. 
The goal is to produce edited views $\{E_i\}_{i=1}^N$ such that (i) each $E_i$ is a faithful edit of $I_i$ according to $y$, and (ii) $\{E_i\}$ are multi-view consistent, i.e. there exists a underlying 3D scene representation $\mathcal{S}$ whose renderings under poses $\{\pi_i\}$ yield $\{E_i\}$.

\subsection{Score Distillation Sampling}

Originally introduced in \emph{DreamFusion}~\cite{poole2022dreamfusion} for 3D generation using 2D diffusion models, 
\textbf{Score Distillation Sampling (SDS)} is an iterative technique for utilizing the generative prior of a pre-trained diffusion model $\epsilon_\psi$ (\emph{teacher}) to tune the parameters $\theta$ of a differentiable neural scene representation $\Phi_\theta$ (\emph{student}). 
At each iteration, the student is queried (rendered) through a differentiable operator $g$, yielding $\hat{\chi}_0 = g(\Phi_\theta)$.
\begin{wrapfigure}[16]{r}{0.13\textwidth}
    \centering
    \includegraphics[width=\linewidth]{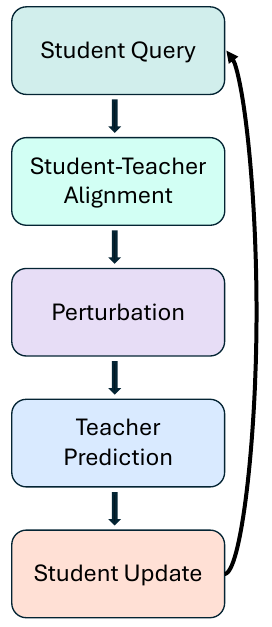}
    \caption{The five SDS stages.}
\end{wrapfigure}
This prediction is then critiqued by the teacher, and the process repeats iteratively, updating $\theta$ until the student encodes a scene representation $\Phi_\theta$ that yields plausible renderings.

The overall SDS framework, can be summarized as a five-stage pipeline, schematically shown in the inset figure:

\textbf{1.~Student Query.}  
The student $\Phi_\theta$ produces an image or latent $\hat{\chi}_0 = g(\Phi_\theta)$ to be critiqued. Commonly  this is a differentiable rendering from a NeRF.  

\textbf{2.~Student-Teacher Alignment.}  
The student output $\hat{\chi}_0$ is mapped to the teacher's input space as $\hat{x}_0$, e.g., through an image encoder. 

\textbf{3.~Perturbation.}  
The aligned prediction $\hat{x}_0$ is perturbed according to the teacher’s forward diffusion process:
$
\hat{x}_t = \alpha_t \hat{x}_0 + \sigma_t \epsilon,
\quad \epsilon \sim \mathcal{N}(0, I). 
$

\textbf{4.~Teacher Prediction.}  
The teacher model $\epsilon_\psi$ processes $\hat{x}_t$ (conditioned on an embedding $y$ and the timestep $t$), and predicts the corresponding noise $  \epsilon_\psi(\hat{x}_t; y, t)$.

\textbf{5.~Student Update.}  
The residual between the sampled noise $\epsilon$ and the teacher's prediction defines the SDS gradient:
\(
\nabla_{\theta} \mathcal{L}_{\mathrm{SDS}} = \mathbb{E}_{t, \epsilon} \left[ w(t)\, \big( \epsilon_\psi(\hat{x}_t; y, t) - \epsilon \big) \frac{\partial \hat{x}_0}{\partial \theta} \right],
\)
where $w(t)$ is a time-dependent weighting term. The gradient is backpropagated to update the student parameters $\theta$.  

SDS variants differ primarily in the specific design choice at each stage.

\subsection{Consistent Sparse-View Editing Through Student Personalization}

In our framework, where the student is a diffusion-based multi-view synthesis model, key SDS stages require specialized adaptations, which we detail in this subsection. Our proposed approach is illustrated in Figure \ref{fig:method}.

\paragraph{Step 1: Student Query.}  
In traditional SDS with NeRFs or 3DGS, the student prediction $\hat{\chi}_0 = g(\Phi_\theta)$ is obtained via differentiable rendering. In our setting, the student is a multi-view diffusion model $\epsilon_\theta$, so the analogue is generating a sample via its denoising  trajectory~\cite{ho2020denoising,song2020denoising}.
Running a full sampling trajectory at each SDS iteration is however slow and computationally expensive, requiring backpropagation through many denoiser evaluations. Instead, we distill incrementally at each student timestep $\tau$, starting from $\tau=\mathrm{T}$ with latents sampled from the Gaussian distribution with student-scheduler specified variance $\{\zeta_\mathrm{T}^i\}_{i=1}^N \sim \mathcal{N}(0, \sigma_{\text{S}}^2 I)$. We compute \emph{single-step predictions} of the clean latents via the Tweedie formula \cite{efron2011tweedie}. These estimates $\{\hat{\zeta}_0^i(\tau)\}$ serve as intermediate student predictions to be critiqued by the teacher, shaping the student’s backward trajectory step by step.

\vspace{-1em}
\paragraph{Step 2: Student-Teacher Alignment.}
Although both student and teacher are latent diffusion models, they operate in different latent spaces and dimensions. A naive approach would decode the student’s predictions $\{\hat{\zeta}_0^i\}$ with its decoder $\mathcal{D}_\text{S}$ and encode them with the teacher encoder $\mathcal{E}_\text{T}$ before adding noise via the teacher’s forward process. However,
 backpropagating through both $\mathcal{D}_\text{S}$ and $\mathcal{E}_\text{T}$ would be  prohibitively expensive.  
Inspired by prior work on convergent representations \cite{asperti2023comparing, lenc2015understanding, huh2024platonic, li2015convergent}, 
which suggests that simple mappings can often bridge the representation spaces of different networks,
we instead resize the student’s latents to the teacher’s expected dimensions $(H_\mathrm{T}, W_\mathrm{T})$ via bilinear interpolation:
$
    \hat{z}_0^i = \mathcal{I}_{bilinear} (\hat{\zeta}_0^i; H_\mathrm{T}, W_\mathrm{T}).
$
For our chosen models, this lightweight approach suffices, suggesting that the student latents implicitly align with the teacher's latent space
during fine-tuning.

\vspace{-1em}
\paragraph{Step 3: Perturbation.}  
The mapped latents are perturbed using the teacher’s forward process
$
    \hat{z}_t^i = \alpha_t \hat{z}_0^i + \sigma_t \epsilon_i,\;
    \epsilon_i \sim \mathcal{N}(0,I),
$
yielding noisy latents $\{\hat{z}_t^i\}$.  
A key design choice is the teacher timestep $t$.  
In standard SDS~\cite{poole2022dreamfusion}, $t$ is drawn uniformly from $[0.02,0.98]$, avoiding extreme noise levels for numerical stability.  This is ill-suited to our setting: early student outputs (large $\tau$) lie off the natural image manifold, so at low $t$ values their diffused versions fall outside the teacher’s distribution, causing unstable guidance.  

Annealed $t$ schedules have also been explored~\cite{huang2023dreamtime, lukoianov2024score}, and a natural variant is to match $t$ with the student timestep $\tau$. Yet this is too restrictive—when $\tau$ is small, forcing $t \approx \tau$ limits the teacher's ability to provide corrective gradients.  We instead use a stochastic schedule:
\(
t \sim \text{TruncNorm}\!\left(\mu=b,\;\sigma=\tfrac{b-\tau}{f},\;a=\tau,\;b=0.95\right),
\)
where $f$ controls skewness. Larger $f$ concentrates probability near $b$, making it more likely for the teacher to operate at higher noise levels. The randomness ensures that the teacher provides strong gradients every few iterations, which we find highly effective for avoiding collapse to poor local minima.  See Appendix \ref{sec:teacher_forward} for further details and visualizations.

\paragraph{Step 4: Teacher Prediction.}

A straightforward application of our framework would pass the perturbed latents 
$\{\hat{z}_t^i\}_{i=1}^N$ as a batch to the monocular teacher U-Net $\epsilon_\psi$, which would then produce independent noise estimates for each latent. Backpropagating such conflicting signals into the student can weaken its multi-view prior, yielding inconsistent final edits. To address this, we introduce a lightweight Random Cross-View Attention (RCVAttn) mechanism that encourages the teacher to generate more consistent edits within each batch. Inspired by attention-based alignment work \cite{khachatryan2023text2video}, we randomly select a \emph{key frame} index $\kappa \sim U\{1,\dots,N\}$ at each iteration. Each frame $i$ attends to the tokens of the key frame:
\begin{equation}
    \text{RCVAttn}(Q, K, V, i) 
    = \text{softmax}\!\left(\frac{Q_i K_\kappa^\top}{\sqrt{d}}\right)V_\kappa,
\end{equation}

\begin{figure}
    \centering
    \includegraphics[width=\linewidth]{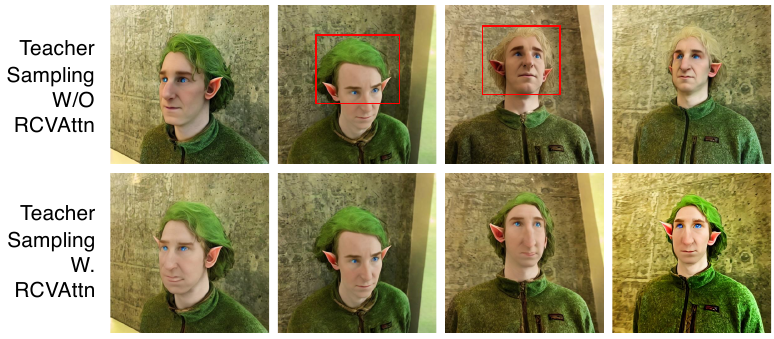}
    \caption{Random Cross-View Attention effect when used for full teacher sampling.}
    \label{fig:rcvattn_ablation}
\end{figure}

where $d$ is the query/key dimensionality. Aligning all frames to query the key frame improves consistency substantially, aiding in retaining the student’s multi-view prior. Unlike expensive extended-attention methods~\cite{wu2023tune, chen2024dge, geyer2023tokenflow}, RCVAttn adds no computational overhead. While non-key frames may experience reduced quality, randomly selecting $\kappa$ ensures all frames occasionally serve as the key, preventing noticeable degradation. The effect of RCVAttn, when applied to full teacher sampling process,  is shown in Fig. \ref{fig:rcvattn_ablation}.

\vspace{-1em}
\paragraph{Step 5: Student Update.}
Finally, the difference between the sampled noise and the teacher’s prediction defines the guidance direction in the SDS objective:
\begin{equation}
    \nabla_{\theta} \mathcal{L}_{\text{SDS}} 
    = \frac{1}{N} \sum_{i=1}^{N} 
    \big( \epsilon_\psi(\hat{z}_t^i; y, I_i, t) - \epsilon_i \big) 
    \frac{\partial \hat{z}_0^i}{\partial \theta} ,
    \label{eq:avg_sds}
\end{equation}
which is backpropagated to update the student  weights.

This completes a single distillation iteration at student timestep $\tau$. We start at $\tau=\mathrm{T}$ and repeat this process for $k$ iterations, personalizing the student model at this timestep to the indented edit. The student then performs a sampling step with its scheduler, producing latents $\{\zeta_{\tau - \Delta\tau}^i\}$, where $\Delta\tau$ is the step size. Distillation resumes at $\tau - \Delta\tau$, repeating $k$ updates before the next sampling step. This nested procedure continues until $\tau = 0$, yielding final edited views that are instruction-faithful and multi-view consistent.

\begin{algorithm}[t]
\caption{\algname}
\label{alg:im2m}
\small
\begin{algorithmic}[1]
\State \textbf{Input:} 
\State \quad $\{I_i\}_{i=1}^N$, $\{\pi_i\}_{i=1}^N$, $y$ \Comment{Images, poses, and text prompt}
\State \quad $\epsilon_\psi$, $\epsilon_\theta$ \Comment{Frozen teacher and trainable student}
\State $z_\text{ref} \gets \mathcal{E}_S(\text{TeacherEdit}(I_\text{ref}, y))$
\State Initialize $\zeta_T^i \sim \mathcal{N}(0, \sigma_S^2 I)$
\For{$\tau = T, T-\Delta\tau, \dots, 0$}
    \For{$k$ steps}
        \State $\{\hat{\zeta}_0^i\} \gets \epsilon_\theta(\{\zeta_\tau^i\}, z_\text{ref}, \{\pi_i\})$
        \State $\hat{z}_0^i \gets \mathcal{I}_\text{bilinear}(\hat{\zeta}_0^i)$
        \State $t \sim \text{TruncNorm}(\mu{=}b,\sigma{=}\frac{b{-}\tau}{f},a{=}\tau,b{=}0.95)$
        \State $\epsilon_i \sim \mathcal{N}(0, I)$
        \State $\hat{z}_t^i \gets \alpha_t \hat{z}_0^i + \sigma_t \epsilon_i$
        \State $\kappa \sim U\{1,\dots,N\}$ \Comment{Select keyframe}
        \State $\{\tilde{\epsilon}_i\} \gets \epsilon_\psi(\{\hat{z}_t^i\}; y, \{I_i\}, t)$ \Comment{With RCVAttn}
        \State $\nabla_\theta \mathcal{L}_\text{SDS} \gets \frac{1}{N} \sum_i (\tilde{\epsilon}_i - \epsilon_i) \frac{\partial \hat{z}_0^i}{\partial \theta}$
        \State $\theta \gets \text{OptimizerStep}(\theta, \nabla_\theta \mathcal{L}_\text{SDS})$
    \EndFor
    \State $\{\zeta_{\tau-\Delta\tau}^i\} \gets \text{StudentStep}(\{\zeta_\tau^i\}, z_\text{ref}, \{\pi_i\}, \Delta\tau)$
\EndFor
\State \textbf{Output:} $E_i \gets \mathcal{D}_T(\hat{z}_0^i)$
\end{algorithmic}
\end{algorithm}

\paragraph{Initialization.} 
Our student model, SEVA, is an "$M$ in, $N$ out" model with $M \geq 1$, meaning that the denoiser $\epsilon_\theta$ requires at least one clean input latent in addition to the $N$ noisy latents.  
As a preprocessing step, we randomly select one of the input frames $I_{\text{ref}} \in \{I_i\}$ and pass it through the 2D teacher editor to generate a valid reference edit $E_{\text{ref}}$.  
This edit is then encoded using SEVA's frozen encoder to obtain a reference latent $z_{\text{ref}} = \mathcal{E}_\text{S}(E_{\text{ref}})$, which serves as the input frame to the denoiser in all distillation iterations. 
The framework is summarized in Algorithm \ref{alg:im2m}.

\section{Experiments}
\label{sec:experiments}

\textbf{Methods in comparison.} We compare with four widely used, open-source methods that also employ InstructPix2Pix as the 2D editor, covering distinct paradigms for multi-view editing: \emph{Instruct-NeRF2NeRF (I-N2N)}~\cite{haque2023instruct} and its 3DGS variant \emph{Instruct-GS2GS (I-GS2GS)}~\cite{vachha2024instruct} (both following the Iterative Dataset Update paradigm), \emph{Text2Video-Zero (T2VZ)}~\cite{khachatryan2023text2video} (a zero-shot image-to-video adaptation), and \emph{DGE}~\cite{chen2024dge} (extended attention for multi-view editing with 3DGS-based consolidation). Since I-N2N requires a trained NeRF, we optimize a Nerfacto~\cite{tancik2023nerfstudio} model on the $N$ input views; similarly, because I-GS2GS and DGE require a 3DGS, we optimize a Splatfacto~\cite{tancik2023nerfstudio} model. All baselines are run with default settings from the official implementations or papers.  

\textbf{Evaluation.} We evaluate our method on scenes from several datasets: I-N2N~\cite{haque2023instruct}, Tanks and Temples~\cite{Knapitsch2017}, CO3D~\cite{reizenstein2021common}, and Mip-NeRF 360~\cite{barron2022mipnerf360}.  
Following prior protocols, for comparison with baselines we apply 20 edits to three standard test scenes from I-N2N (full edit set detailed in Appendix~\ref{sec:evaluation_edits}); qualitative results on additional scenes appear in Appendix~\ref{sec:results_for_additional_scenes}.  
Per-frame edit quality and cross-view consistency are assessed with three CLIP-based~\cite{radford2021learning} metrics commonly used in prior work~\cite{haque2023instruct, chi2025disco3d, chen2024dge}:  
(i) \emph{CLIP Similarity}, the cosine similarity between an edited image and the prompt;  
(ii) \emph{CLIP Directional Similarity}~\cite{gal2022stylegan, brooks2023instructpix2pix}, which measures alignment between prompt change and image change;
(iii) \emph{CLIP Directional Consistency}~\cite{haque2023instruct}, which quantifies multi-view consistency by comparing the relative changes between pairs of original views $O_i, O_j$ and their corresponding edited views $E_i, E_j$ via
\(
cos\_sim\big( \phi(O_i) - \phi(O_j), \; \phi(E_i) - \phi(E_j) \big),
\)
where $\phi(\cdot)$ denotes the CLIP embedding. This metric captures whether the semantic difference between two views is preserved after editing. Unlike the original formulation, which considers only consecutive frames, we average over all $\binom{N}{2}$ pairs to account for our unordered, sparse-view setting.  

We use $N=4$ frames in main experiments, with additional results for larger $N$ in Appendix~\ref{sec:results_more_frames}. Full implementation details, are detailed in Appendix~\ref{sec:implementation_details}.

\subsection{Comparison with Prior Work}

\begin{figure*}[t]
\includegraphics[width=0.99\linewidth]{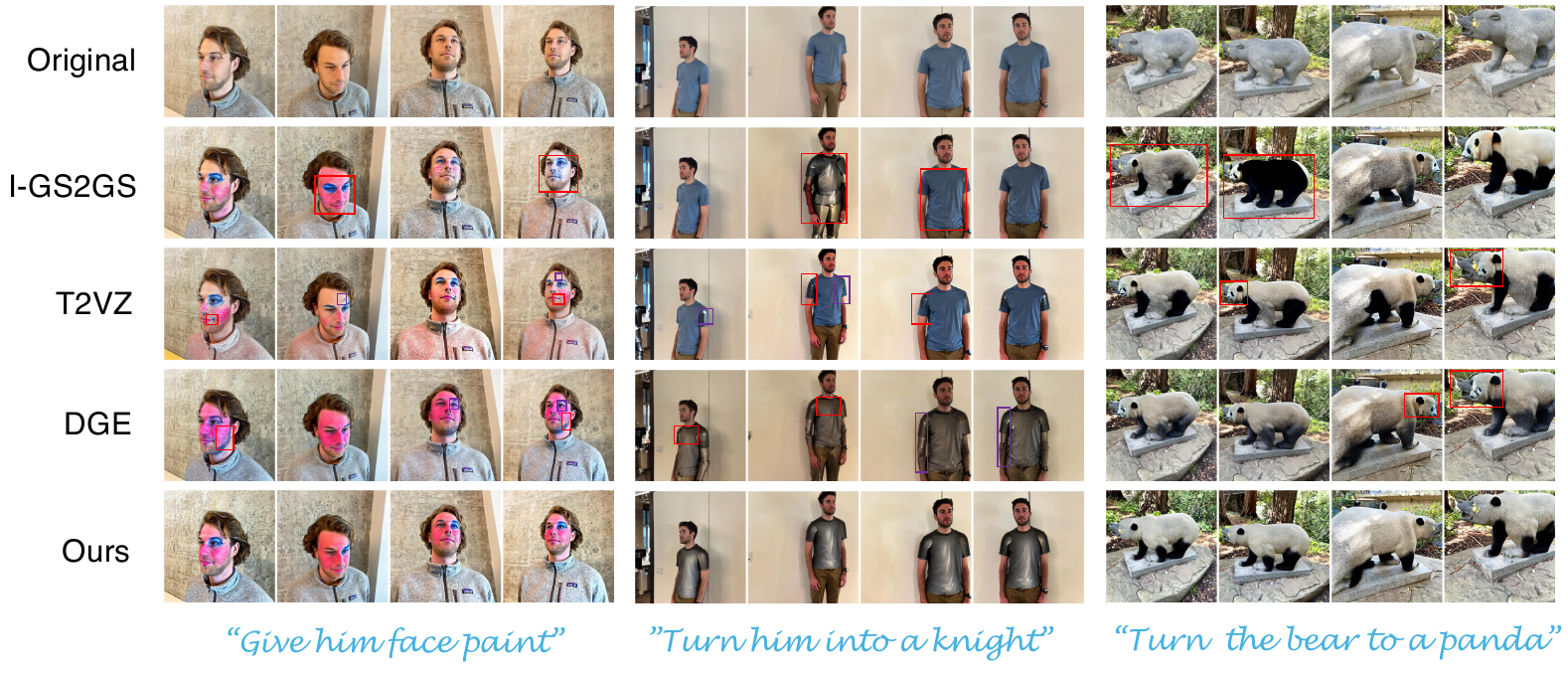}

\includegraphics[width=0.99\linewidth]{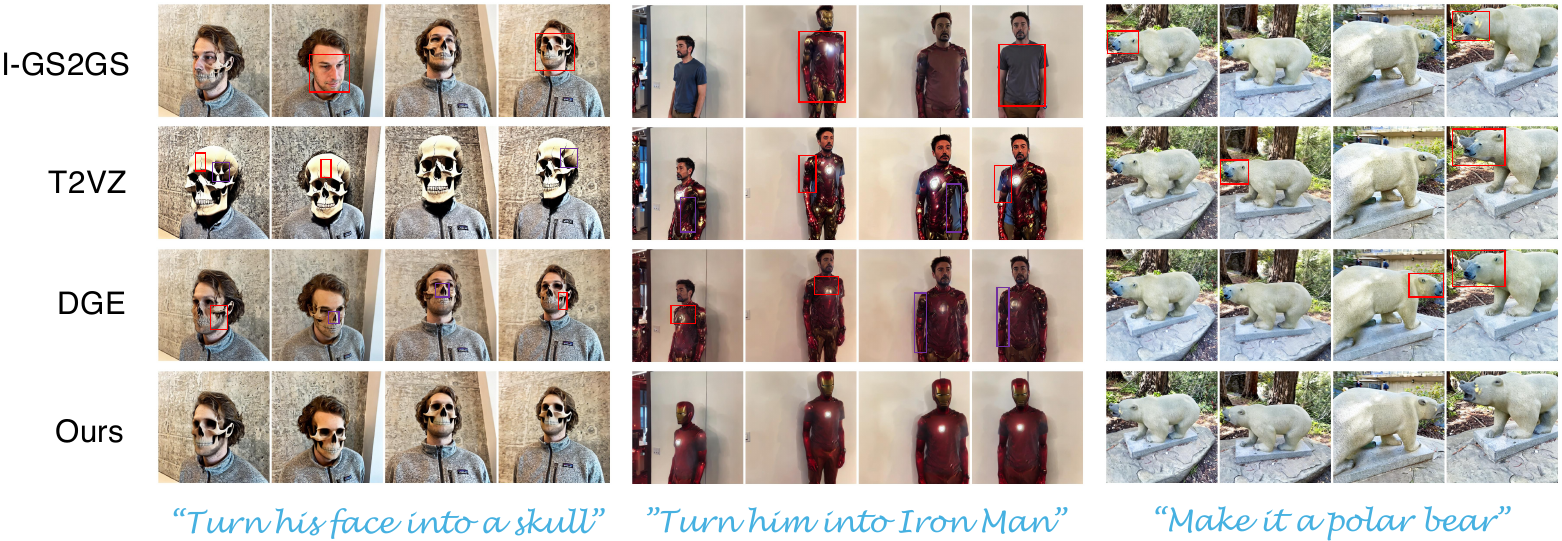} 

    \caption{Qualitative comparison with prior work. The top row shows the original scenes, and the lower rows present edits from different methods. Matching red or purple rectangles indicate pairs of inconsistent regions, which frequently appear in baselines but not in our edits. Please zoom in electronically for details; enlarged views are provided in Appendix \ref{sec:extended_comparison_to_baselines}.}

    \label{fig:comparison_to_baselines}
\end{figure*}

\begin{table}[t]
    \centering
    \setlength{\tabcolsep}{3.8pt} 
    \renewcommand{\arraystretch}{1.1} 
    \begin{tabular}{lccc}
        \toprule
        \textbf{Method} & \textbf{CLIP Cons. $\uparrow$} & \textbf{CLIP Sim. $\uparrow$} & \textbf{CLIP Dir. $\uparrow$} \\
        \midrule
        I-N2N & 0.034 & 0.196 & 0.105 \\
        I-GS2GS & 0.314 & 0.253 & 0.169 \\
        T2VZ & 0.310 & 0.251 & 0.159 \\
        DGE & 0.287 & 0.256 & \textbf{0.182} \\
        \textbf{Ours} & \textbf{0.342} & \textbf{0.258} & 0.173 \\
        \bottomrule
    \end{tabular}
    \caption{Comparison of methods across view consistency, semantic alignment, and edit performance.}
    \label{tab:comparison_to_baselines}
\end{table}

Quantitative results are reported in Table~\ref{tab:comparison_to_baselines}, and representative qualitative comparisons are shown in Figure~\ref{fig:comparison_to_baselines}. Enlarged visualizations and additional examples are included in Appendix \ref{sec:extended_comparison_to_baselines}.  

Our method achieves the highest performance in \textit{CLIP Directional Consistency (CLIP Cons.)}, indicating that edits remain more consistent across different views. Importantly, this does not come at the cost of per-frame edit quality, as demonstrated by CLIP Sim. and CLIP Dir. scores, where our method is either superior to or competitive with the baselines.  

Notably, I-N2N fails completely in this sparse-view setting. We observed that Nerfacto struggles to fit the scene, producing severe floater artifacts even when rendering source poses. These distortions lie out of distribution for the 2D editor, leading to unusable edits as shown in Appendix \ref{sec:in2n_limitations}.

The advantages of our approach compared to other baselines, are most clearly demonstrated qualitatively. 
In the sparse-view setting, baseline methods often struggle to maintain consistent edits across viewpoints due to two factors: 
(i)~3DGS-based consolidation becomes unreliable with limited views, as the 3DGS tends to overfit the training images \cite{zhang2024cor, zhu2024fsgs}; and 
(ii)~cross-frame attention, while improving general appearance alignment, fails to enforce fine-grained consistency. 
Figure~\ref{fig:comparison_to_baselines} illustrates these issues: I-GS2GS edits remain largely view-independent (e.g., the \emph{Face Paint} edit), while T2VZ and DGE, though producing roughly appearance-consistent edits, introduce inconsistency in the details—such as mismatched sleeve and chest textures in the \emph{Knight} and \emph{Iron Man} edits, varying face paint colors and intensities in \emph{Face Paint}, and view-dependent differences in \emph{Skull} details such as the nose, cheek, and forehead. 
The lack of robust 3D consistency is especially evident in the \emph{Bear} edit, which exhibits Janus-like multi-face artifacts.
In contrast, \algname\ produces edits that are not only faithful to the instruction but also highly consistent across all views, without sacrificing image quality. This combination of instruction alignment, visual fidelity, and strong 3D consistency represents a clear improvement over existing approaches in the sparse-view setting.

\begin{table}[t]
    \centering
    \setlength{\tabcolsep}{3.5pt} 
    \renewcommand{\arraystretch}{1.1} 
    \begin{tabular}{lcccc}
        \toprule
        \textbf{Method} &
        \begin{tabular}[c]{@{}c@{}}\textbf{\# Inconsist.}\\[-2pt]$\downarrow$\end{tabular} &
        \begin{tabular}[c]{@{}c@{}}\textbf{Scene}\\[-2pt]\textbf{Win \%} $\uparrow$\end{tabular} &
        \begin{tabular}[c]{@{}c@{}}\textbf{Consist.}\\[-2pt]\textbf{\%} $\uparrow$\end{tabular} &
        \begin{tabular}[c]{@{}c@{}}\textbf{Inconsist.}\\[-2pt]\textbf{\%} $\downarrow$\end{tabular} \\
        \midrule
        DGE & 2.02 & 25.0 & 34.0 & 31.0 \\
        \textbf{Ours} & \textbf{1.34} & \textbf{75.0} & \textbf{65.0} & \textbf{13.0} \\
        \bottomrule
    \end{tabular}
    \caption{Human study of multi-view consistency. Differences are statistically significant; full methodology appear in Appendix~\ref{sec:3d_consistency_survey}.}
    \label{tab:mv3d_human}
\end{table}

\paragraph{Human survey.} To further validate the 3D consistency advantage of \algname, we conducted a human survey comparing it to the strongest baseline, DGE. Raters were asked to identify and mark inconsistencies in the multi-view edited scenes produced by both algorithms. As shown in Table~\ref{tab:mv3d_human}, \algname\ significantly outperforms DGE by producing fewer inconsistencies on average, achieving higher scene win percentages, and generating a greater proportion of "consistent" scenes (with 1 or fewer inconsistencies). Additionally, \algname\ results in far fewer "inconsistent" scenes (with 3 or more inconsistencies), demonstrating its superior multi-view consistency. See Appendix \ref{sec:3d_consistency_survey} for full details of the methodology and statistical significance.

\begin{table}[t]
    \centering
    \setlength{\tabcolsep}{3.4pt}
    \renewcommand{\arraystretch}{1.1} 
    \small
    \begin{tabular}{llccc}
        \toprule
        \textbf{SDS Stage} & \textbf{Config} &
        \begin{tabular}[c]{@{}c@{}}\textbf{CLIP}\\[-2pt]\textbf{Cons.} $\uparrow$\end{tabular} &
        \begin{tabular}[c]{@{}c@{}}\textbf{CLIP}\\[-2pt]\textbf{Sim.} $\uparrow$\end{tabular} &
        \begin{tabular}[c]{@{}c@{}}\textbf{CLIP}\\[-2pt]\textbf{Dir.} $\uparrow$\end{tabular} \\
        \midrule
        \NA & Student Only & \textcolor{red}{0.014} & \textcolor{red}{0.212} & 0.161 \\
        \NA & Teacher Only & \textcolor{red}{0.228} & 0.252 & 0.184 \\
        \midrule
        Initialization (0) & Source ref. Frame & 0.326 & 0.264 & 0.174 \\ 
        \midrule
        Alignment (2) & Learned Mapping & 0.287 & 0.259 & 0.180 \\
        \midrule
        \multirow{2}{*}{Perturbation (3)} 
            & Uniform $t$ & 0.363 & 0.260 & \textcolor{red}{0.146} \\
            & $\tau$-matched $t$ & 0.435 & 0.231 & \textcolor{red}{0.107} \\
        \midrule
        Teacher Pred. (4) & W/O RCVAttn & \textcolor{red}{0.230} & 0.260 & 0.175 \\
        \midrule
        \multicolumn{2}{c}{\textbf{Full}} & 0.337 & 0.263 & 0.178 \\
        \bottomrule
    \end{tabular}
    \caption{Ablation study evaluating different design choices. Weak results are highlighted in red.}
    \label{tab:ablation_study}
\end{table}

\subsection{Ablation Study}
\label{sec:ablation}

We conduct an ablation study on 6 representative edits (listed in Appendix \ref{sec:evaluation_edits}) to assess the contributions of our design choices across the SDS pipeline. Quantitative results are summarized in Table~\ref{tab:ablation_study}, with particularly weak results highlighted in red. Our findings show that each component is essential to achieve both faithful edits and strong multi-view consistency.  

\textbf{Role of teacher and student.}  
We first test the student and teacher models in isolation. In the \emph{Student Only} setting, the student is given an edited frame as input and asked to sample additional views. This fails for several reasons: the student never sees the scene content captured by the other frames, leading to poor faithfulness; the SEVA prior struggles under single-view input; and we suspect that the edited scenes lie out-of-distribution for the model. This suggests that our approach \emph{distills new capabilities into the student}, rather than simply searching within its existing sampling distribution. Conversely, in the \emph{Teacher Only} setting, we rely on the teacher to edit each view independently. While individual frames adhere to the instruction edit, the lack of a 3D prior leads to severe cross-view inconsistency, as reflected by the low CLIP Consistency score. We present representative visualizations in Appendix \ref{sec:limitations_teacher_student}.
We further ask whether a lightweight coupling could suffice: we adapt SDEdit \cite{meng2021sdedit} to our setting, using the teacher’s per-view edits as a coarse initialization for the multi-view student (Appendix~\ref{sec:sdedit}). In practice, this naïve fusion yields low-quality, view-inconsistent results.
Together, these results confirm the necessity of distilling from the teacher into the student, rather than using either in isolation.  

\textbf{Initialization stage.}  
In the \emph{Source ref. Frame} setting, we input one of the original frames to the student encoder, to serve as the reference latent, without editing it first: $z_{\text{ref}} = \mathcal{E}_\text{S}(I_{\text{ref}})$. Skipping the reference frame edit negatively affected the distillation process, as the initial student predictions are further away from the target. This results in slightly lower  multi-view consistency.

\begin{figure}
    \centering
    \includegraphics[width=0.7\linewidth]{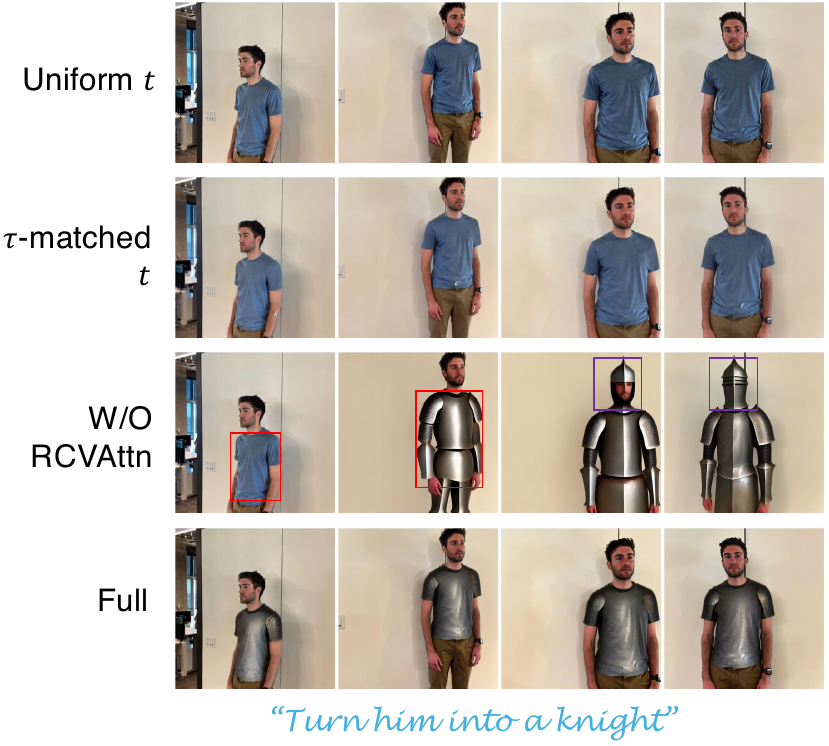}
    \caption{Failure cases from variants of the perturbation and teacher prediction stages.  
Rows 1–2: alternative forward schedules  collapse to near-identity edits.  
Row 3: removing RCVAttn breaks multi-view coherence.  
Row 4: full method output.}

    \label{fig:perturnation_ablation}
\end{figure}

\textbf{Alignment stage.}  
Following findings in prior work on latent space alignment \cite{huh2024platonic, li2015convergent}, we replaced the bilinear interpolation with a learnable convolutional mapping (\emph{Learned Mapping}), optimized during distillation, but found this brought no measurable benefit -- the necessary transformation is likely captured during the fine-tuning of the student.

\textbf{Perturbation stage.}  
We evaluated two alternatives to our proposed forward schedule: \emph{Uniform $t$} (similar to \citet{poole2022dreamfusion}), where $t$ is sampled uniformly in $[0.05, 0.95]$, and \emph{$\tau$-matched $t$}, where the teacher’s timestep follows the student’s. 
Both variants tend to collapse to near-identity reconstructions of the input scene, which explains their paradoxically high CLIP Consistency: such outputs are trivially consistent but fail to realize the intended edit as reflected in their low CLIP Directional scores. Visual examples are provided in the first two rows of Figure \ref{fig:perturnation_ablation}.

\textbf{Teacher prediction stage.}  
Disabling our Random Cross-View Attention mechanism (\emph{W/O RCVAttn}) leads the teacher to process each perturbed latent independently. Without cross-view coupling, the student receives conflicting signals across views, leading to degraded multi-view consistency and breaking its 3D prior. This is again reflected in low CLIP Consistency, and illustrated in the third row of Figure \ref{fig:perturnation_ablation}.

\begin{figure}
\centering
\includegraphics[width=\linewidth]{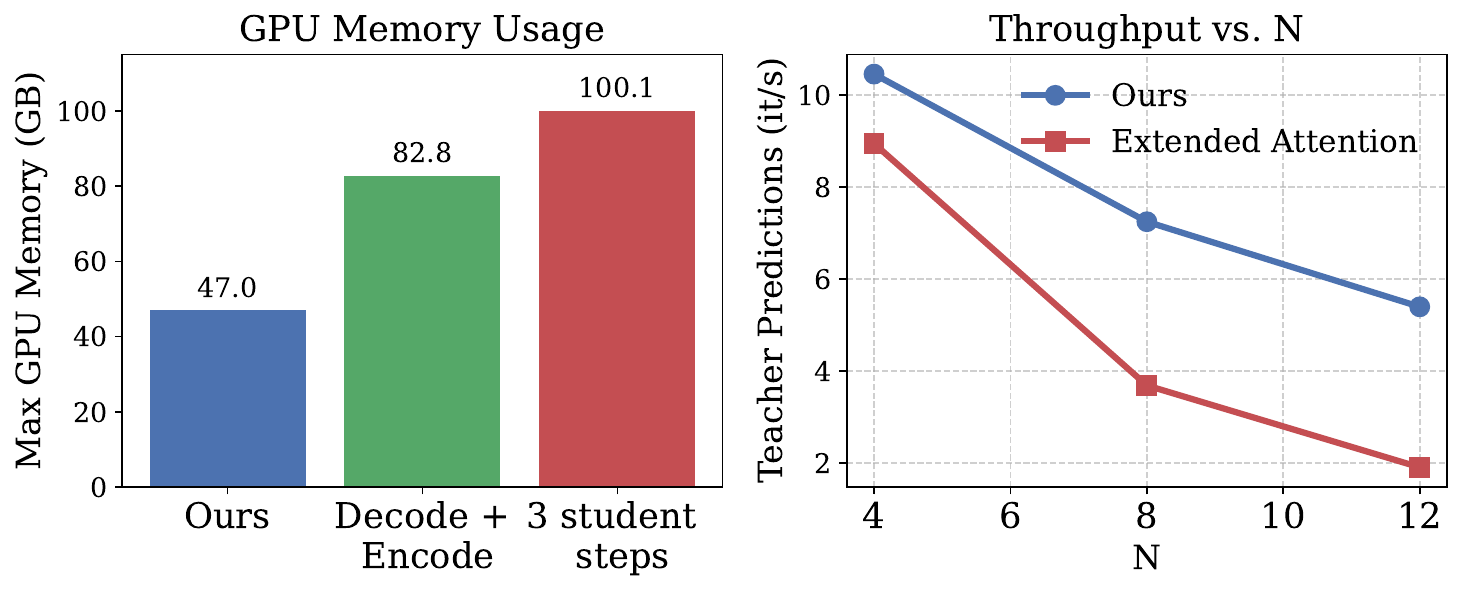}
\caption{Efficiency analysis.  
Left: peak GPU memory usage for alternative student update and alignment strategies.  
Right: throughput degradation with extended attention as the number of views $N$ increases.}

    \label{fig:efficiency}
\end{figure}

\textbf{Efficiency considerations.}  
Several components of \algname\ were explicitly designed for memory and compute efficiency. Our choice of using a single-step prediction in the \emph{Student Query} stage is crucial: a three-step alternative more than doubles peak memory usage for $N=4$ views (Figure \ref{fig:efficiency}, left). 
In the \emph{Student-Teacher Alignment} stage, interpolating latents rather than passing through the student’s decoder and teacher’s encoder reduces memory usage by over $40\%$. Finally, replacing the RCVAttn module with full extended attention significantly degraded throughput, worsening as the number of frames increased (Figure \ref{fig:efficiency}, right). We additionally experimented with fine-tuning the student using LoRA~\cite{hu2022lora} rather than updating the full U-Net. While more parameter-efficient, this approach underperformed, and we leave the adaptation of lightweight variants to future work.

\subsection{Beyond Image Editing}\label{sec:beyond_edit}
Our approach is not inherently limited to  editing tasks. In principle, any pre-trained image-to-image diffusion model can serve as the teacher, with the multi-view student acting as a consolidator to produce a multi-view–to–multi-view solution. To explore this, we experimented with multi-view conditional generation. Specifically, we employed pre-trained ControlNets \cite{zhang2023adding} as teachers to translate multiple depth maps or Canny edge maps of a 3D scene into consistent RGB images. Qualitative examples are provided in Appendix \ref{sec:results_beyond_editing}. While the outputs were faithful to the conditioning inputs and maintained multi-view consistency, they often exhibited excessive blurriness—a known artifact of SDS-based optimization \cite{poole2022dreamfusion}.

\section{Discussion: Parallel to Diffusion Guidance}

In standard diffusion guidance \cite{bansal2023universal, chung2022diffusion}, the model predictions at a given timestep are critiqued, and the resulting gradient is used to modify the sampling trajectory.  
In our framework, rather than applying such potentially unstable updates to the latents, we backpropagate the guidance signal to the \emph{student’s weights}. This approach effectively transfers the teacher’s knowledge without making aggressive modifications to the latents themselves, avoiding divergence from the target distribution.

\section{Conclusion, Limitations, and Future Work}
We presented \algname, a novel framework for multi-view image editing that achieves high multi-view consistency in sparse-view settings, where prior methods typically fail. While effective, our approach inherits the limitations of its backbones, specifically InstructPix2Pix and SEVA, which can struggle with certain edit prompts or with maintaining perfect consistency across views. Given our framework's modular nature, we anticipate that integrating stronger future backbones could mitigate these issues.
Additionally, \algname requires multiple distillation iterations per noise level, making it more than twice as slow as our strongest competitor, DGE. We plan to explore strategies to reduce this overhead in future work. 
Finally, as discussed in Section~\ref{sec:beyond_edit}, our framework is potentially general and applicable to a range of image manipulation tasks beyond editing.  However, performance on these tasks currently lags behind our editing results, often producing blurry outputs. We leave the investigation of these directions to future work.

\paragraph{Acknowledgments.}
Or Litany acknowledges support from the Israel Science Foundation (grant 624/25) and the Azrieli Foundation Early Career Faculty Fellowship. This research was also supported in part by an academic gift from Meta. The authors gratefully acknowledge this support.

{
    \small
    \bibliographystyle{ieeenat_fullname}
    \bibliography{references}
}

\clearpage
\onecolumn
\addcontentsline{toc}{section}{Appendix Overview}

\begin{center}
    \tableofcontents
\end{center}

\clearpage
\twocolumn
\appendix

\appendix 

\section{Implementation Details}
\label{sec:implementation_details}

We use SEVA 1.1 \citep{zhou2025stable} as the pre-trained student model and InstructPix2Pix \citep{brooks2023instructpix2pix} from the Diffusers library \citep{von_Platen_Diffusers_State-of-the-art_diffusion} as the frozen teacher. Consistent with prior observations \cite{haque2023instruct, chen2024dge}, the teacher’s classifier-free guidance (CFG) scales for both prompt and input image have a significant effect on the \emph{degree of edit intensity}—a factor that is often subjective and a matter of personal taste. For most edits we adopt the default $S_T = 7.5$ for the prompt and $S_I = 1.5$ for the input image, with adjustments detailed appendix \ref{sec:evaluation_edits}.  
We perform distillation over  $40$ student timesteps ($\Delta\tau = \frac{1}{40})$, with $k = 50$ updates per step. Optimization is done with AdamW \citep{loshchilov2017decoupled}, using a maximum learning rate of $1 \times 10^{-4}$ after 200 iterations of linear warm-up, followed by cosine decay down to $5 \times 10^{-5}$. This yields just over 2000 distillation iterations per experiment, which take about 40 minutes on a single NVIDIA H200 GPU. The input frames are selected randomly.

\subsection{Teacher Forward Schedule}
\label{sec:teacher_forward}

\begin{figure}[H]
    \centering
    \includegraphics[width=\linewidth]{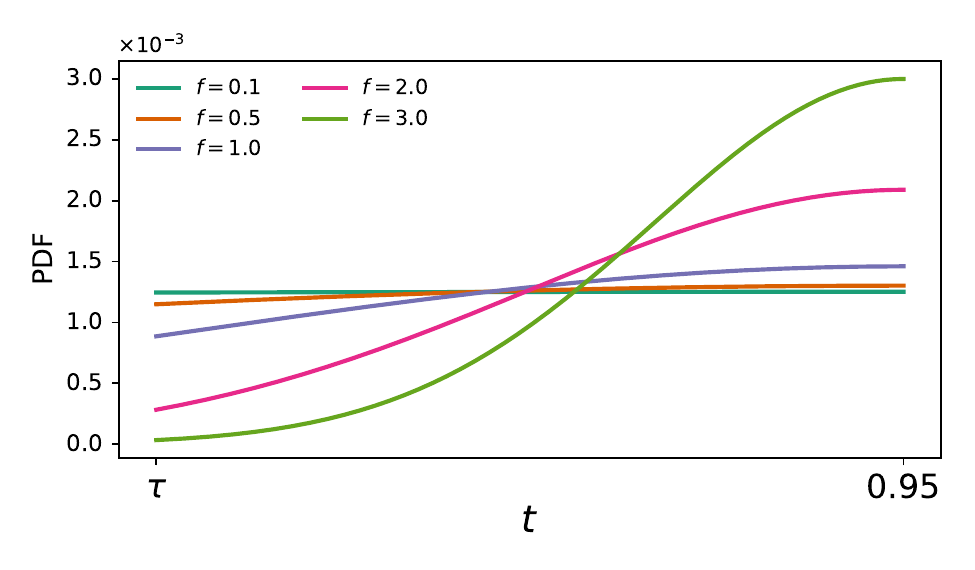}
    \caption{Teacher timestep schedule for different skewness factors $f$.}
    \label{fig:placeholder}
\end{figure}

We employ a stochastic schedule for the teacher forward process, sampling timesteps as
\[
t \sim \text{TruncNorm}\!\big(\mu=b,\;\sigma=(b-\tau)/f,\;a=\tau,\;b=0.95\big),
\]
where $\tau$ is the current student timestep and $f$ controls the skewness of the distribution. Larger $f$ concentrates probability near $b$, making the teacher more likely to operate at higher noise levels. The stochasticity ensures the teacher provides strong gradients every few iterations, which we find effective for avoiding collapse to poor local minima. Figure~\ref{fig:placeholder} illustrates how different $f$ values shape the probability distribution. In practice, we use $f = 0.5$, yielding an approximately uniform distribution over $[\tau, 0.95]$.

\section{3D Consistency Evaluation}

A strong multi-view generative model should produce image sequences in which each frame is individually high-quality and the \emph{entire sequence} is {3D consistent. Unlike image quality or reconstruction accuracy, however, there is currently no widely accepted metric for evaluating 3D consistency. Recent work has begun to address this gap, but existing approaches remain limited.

MEt3R \cite{asim2025met3r} leverages Dust3r \cite{wang2024dust3r} to obtain dense 3D reconstructions for pairs of images in a sequence, and measures feature-space reprojection errors in overlapping regions to assess consistency. While promising, this approach is not well suited to our \emph{sparse-view} setting, where overlapping content between images is minimal. PRISM \cite{stern2025appreciateviewtaskawareevaluation} instead uses diffusion features that capture source--target viewpoint relationships and is, in principle, better suited to large viewpoint changes, but its current implementation is restricted to object-centric scenes.

We also explored using recent feed-forward 3D reconstruction models to lift an evaluated sequence into 3D and measure reprojection error. The underlying assumption is that 3D-consistent scenes should yield lower reprojection error than inconsistent ones. To this end, we employed AnySplat \cite{jiang2025anysplat} as the sparse-view reconstruction model. While its performance is impressive, it proved insufficient for our purposes: even 3D-consistent edited scenes---and occasionally ground-truth scenes---produced high reconstruction error. This may be due both to the inherent difficulty of the sparse-view setting and to edited scenes falling outside the model’s training distribution.

Given these limitations, we adopt the \textbf{CLIP Directional Consistency} metric introduced in prior 3D editing work \cite{haque2023instruct}, as detailed in Section~\ref{sec:experiments}. To complement this automatic metric, we additionally conduct a user study, described in the following subsection.

\subsection{Human Survey}
\label{sec:3d_consistency_survey}

\paragraph{Protocol.}
We compare \algname\ against the strongest baseline, DGE, in a human study focused solely on \emph{multi-view consistency}.
Each rating task shows a set of four images from the same edited scene. Raters are instructed to mark every inconsistency they observe by placing \emph{one pair of rectangles} on the contradictory areas in two views that disagree (each pair counts as one inconsistency). Figure \ref{fig:survey_example} presents an example of the inconsistencies marked by a rater.
Each algorithm produces 20 edited scenes (from Table \ref{tab:editing_scenes}), resulting in 40 total sets of images. On average, each set receives 5 independent ratings, for a total of $200$ rated tasks ($100$ per algorithm).

\paragraph{Primary metric.}
For each rated task we gather the \emph{number of inconsistency pairs} (two rectangles), denoted \#Pairs. Lower is better (fewer inconsistencies).

\paragraph{Aggregate comparisons.}
We report four metrics (Table~\ref{tab:mv3d_human}):
(i) \textbf{\# Inconsistencies}—mean \#Pairs per task;
(ii) \textbf{Scene Win \%}—fraction of scenes where an algorithm’s scene mean \#Pairs is lower than the competitor;
(iii) \textbf{Consistent \%}—fraction of task ratings with $\#\text{Pairs}\leq 1$;
(iv) \textbf{Inconsistent \%}—fraction with $\#\text{Pairs}\geq 3$.

\paragraph{Statistical tests.}
Because per-set sample sizes are small, we avoid normality assumptions and use non-parametric or exact tests:
\begin{itemize}
  \item \textbf{Paired by scene \#Pairs means:} two-sided sign-flip permutation test on per-scene difference of means (20 scenes, 20{,}000 randomizations). Result: $\overline{\Delta}=\text{mean}(\text{\algname} - \text{DGE})=-0.5607$, two-sided $p=0.0371$.
  \item \textbf{Scene wins:} exact binomial sign test on wins. \algname\ wins $15/20$ scenes; one-sided $p=0.020695$ (testing \algname\,$>$\,DGE), two-sided $p=0.041389$.
  \item \textbf{Rates (\,$\#\text{Pairs}\leq 1$ and $\#\text{Pairs}\geq 3$\,):} Fisher’s exact test on $2\times2$ counts (algorithm $\times$ indicator).
  For \emph{Consistent} ($\leq1$): \algname\ $65/100$ vs.\ DGE $34/100$; two-sided $p=1.9\!\times\!10^{-5}$, one-sided ( \algname\,$>$\,DGE ) $p=1.0\!\times\!10^{-5}$.
  For \emph{Inconsistent} ($\geq3$): \algname\ $13/100$ vs.\ DGE $31/100$; two-sided $p=0.003405$, one-sided ( \algname\,$<$\,DGE ) $p=0.001703$.
\end{itemize}

\paragraph{Takeaway.}
Across all analyses, \algname\ exhibits fewer inconsistencies on average, wins on most scenes, substantially more highly consistent ratings ($\leq1$), and far fewer inconsistent ratings ($\geq3$). All reported advantages are statistically significant under the non-parametric / exact tests above.

\section{Evaluation Scenes and Edits}
\label{sec:evaluation_edits}

We detail in Table \ref{tab:editing_scenes} the edits used in our evaluations, applied to the standard Face, Bear, and Person scenes from the Instruct-NeRF2NeRF dataset \cite{haque2023instruct}. The \emph{Edit Prompt} is the editing instruction provided as input to the evaluated methods, while the \emph{Original Prompt} and \emph{Edited Prompt} are employed for CLIP-based evaluation. For each edit, we also report the teacher’s text and image CFG scales, $s_T$ and $s_I$, used in quantitative evaluation. Edits with bolded prompts indicate those selected for the ablation experiments.

\begin{table*}
\centering
\small
\setlength{\tabcolsep}{4pt}
\renewcommand{\arraystretch}{1.2}
\resizebox{\textwidth}{!}{%
\begin{tabular}{|c|p{3.5cm}|p{3.5cm}|p{4.5cm}|c|c|}
\hline
\textbf{Scene} & \textbf{Original Prompt} & \textbf{Edit Prompt} & \textbf{Edited Prompt} & \textbf{Text CFG} & \textbf{Image CFG} \\
\hline

\multirow{8}{*}{\centering Face} 
& \multirow{8}{3.5cm}{\centering "A man with curly hair in a grey jacket"} 
& "Give him a Venetian mask" & "A man with curly hair in a grey jacket with a Venetian mask" & 7.5 & 1.5 \\ \cline{3-6}
& & "Turn him into a vampire" & "A vampire with curly hair" & 7.5 & 1.5 \\ \cline{3-6}
& & \textbf{"Turn him into Tolkien Elf"} & "A Tolkien Elf with curly hair" & 9.0 & 1.5 \\ \cline{3-6}
& & "Turn him into batman" & "A batman" & 7.5 & 1.5 \\ \cline{3-6}
& & \textbf{"Turn his face into a skull"} & "A man with a skull head in a grey jacket" & 7.5 & 1.5 \\ \cline{3-6}
& & "Turn him into Albert Einstein" & "Albert Einstein with curly hair" & 7.5 & 1.5 \\ \cline{3-6}
& & "Turn it to a Van Gogh painting" & "A Van Gogh painting of a man with curly hair in a jacket" & 7.5 & 1.5 \\ \cline{3-6}
& & "Give him face paint" & "A man with curly hair in a grey jacket with face paint" & 7.5 & 1.5 \\ 
\hline

\multirow{4}{*}{\centering Bear} 
& \multirow{4}{3.5cm}{\centering "A stone bear in a garden"}
& \textbf{"Turn the bear to a panda bear"} & "A panda bear in a garden" & 6.0 & 1.5 \\ \cline{3-6}
& & "Turn the bear to a polar bear" & "A polar bear in a garden" & 6.0 & 1.5 \\ \cline{3-6}
& & \textbf{"Turn the bear to a grizzly bear"} & "A grizzly bear in a garden" & 5.5 & 1.5 \\ \cline{3-6}
& & "Turn the bear to a wooden bear" & "A wooden bear in a garden" & 8.5 & 1.5 \\
\hline

\multirow{8}{*}{\centering Person} 
& \multirow{8}{3.5cm}{\centering "A man standing next to a wall wearing a blue T-shirt and brown pants"}
& "Turn him into Iron Man" & "An Iron Man standing next to a wall" & 7.5 & 1.5 \\ \cline{3-6}
& & "Turn the man into a robot" & "A robot standing next to a wall" & 5.5 & 1.8 \\ \cline{3-6}
& & "Make him in a suit" & "A man standing next to a wall wearing a suit" & 6.5 & 1.8 \\ \cline{3-6}
& & \textbf{"Turn him into a clown"} & "A clown standing next to a wall" & 6.0 & 1.8 \\ \cline{3-6}
& & "Make him into a marble statue" & "A marble statue of a man next to a wall" & 7.5 & 1.5 \\ \cline{3-6}
& & "Turn him into a cowboy with a hat" & "A cowboy wearing a hat standing next to a wall" & 6.0 & 1.5 \\ \cline{3-6}
& & "Turn him into a soldier" & "A soldier standing next to a wall" & 7.5 & 1.5 \\ \cline{3-6}
& & \textbf{"Turn him into a knight"} & "A knight standing next to a wall" & 6.0 & 1.5 \\ 
\hline

\end{tabular}%
}
\caption{Prompts and CFG values for each edit used for quantitative evaluation.}
\label{tab:editing_scenes}
\end{table*}

\section{Limitations of Instruct-NeRF2NeRF in Sparse-View Settings}
\label{sec:in2n_limitations}

In the sparse-view regime, Instruct-NeRF2NeRF (I-N2N) \cite{haque2023instruct} fails to produce coherent results. Its underlying Nerfacto \citep{tancik2023nerfstudio} model, trained with default configurations for 30K iterations, struggles to reconstruct the scene accurately, generating severe floater artifacts even when rendering the original input poses. These distortions fall far outside the distribution expected by the 2D editor, rendering the resulting edits unusable. Figure~\ref{fig:in2n-fails} illustrates two representative examples of such failures, corresponding to the \emph{Clown} and \emph{Face Paint} edits.

\section{Student and Teacher Limitations}
\label{sec:limitations_teacher_student}

Figure~\ref{fig:student_teacher_limitations} illustrates the limitations of the student (SEVA) and teacher (Instruct-Pix2Pix) when used individually. Even on the unedited scene, SEVA can struggle to produce high-quality results with only a single input frame, as shown in the second row. When used as an editing baseline—receiving a single edited frame and asked to generate the remaining views—it fails to produce coherent frames (third row). Individual predictions from the teacher (final row) are independent across views, resulting in inconsistent and sometimes implausible edits.

\subsection{SDEdit-Style Fusion Without Distillation}
\label{sec:sdedit}

We test a simple, distillation-free fusion of the 2D editing teacher with the multi-view student by adapting SDEdit \cite{meng2021sdedit} to our setting. 
(1) We first produce per-view edits by independently sampling the teacher. 
(2) Each edited image is mapped into the student’s latent space by decoding with the teacher’s decoder and re-encoding with the student’s encoder. 
(3) We perturb these latents using the student’s forward diffusion process at noise levels \(t\in\{0.25,0.5,0.75\}\). 
(4) Finally, we denoise with the multi-view student, conditioning on the edited view as the input latent (see Section~\ref{sec:method}). 
In principle, the teacher’s edits provide a semantic initialization while the student enforces multi-view consistency.

In practice, this SDEdit-style initialization fails to produce coherent multi-view results: edits are low-quality and inconsistent across views for all tested \(t\) (see Figure~\ref{fig:sdedit_limitations}). 
This ablation underscores the need for explicit distillation, as implemented in \algname.

\begin{figure*}[t]
\centering
\includegraphics[width=0.95\linewidth]{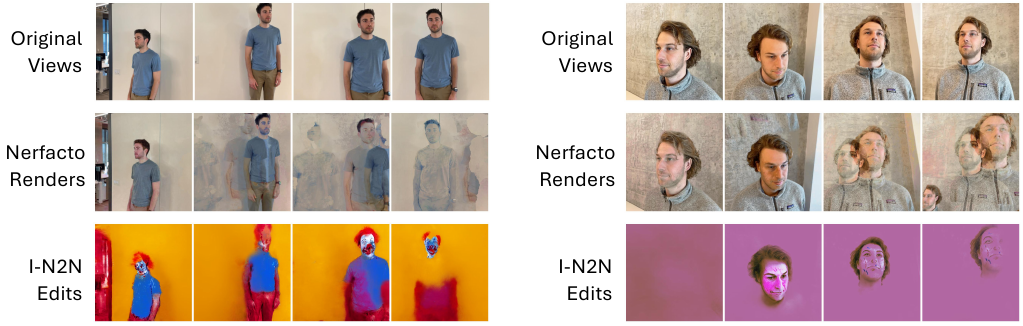}
\caption{Examples for I-N2N failures in the sparse-view setting.}
\label{fig:in2n-fails}
\vspace{1em}

\begin{minipage}{0.48\linewidth}
    \centering
    \includegraphics[width=\linewidth]{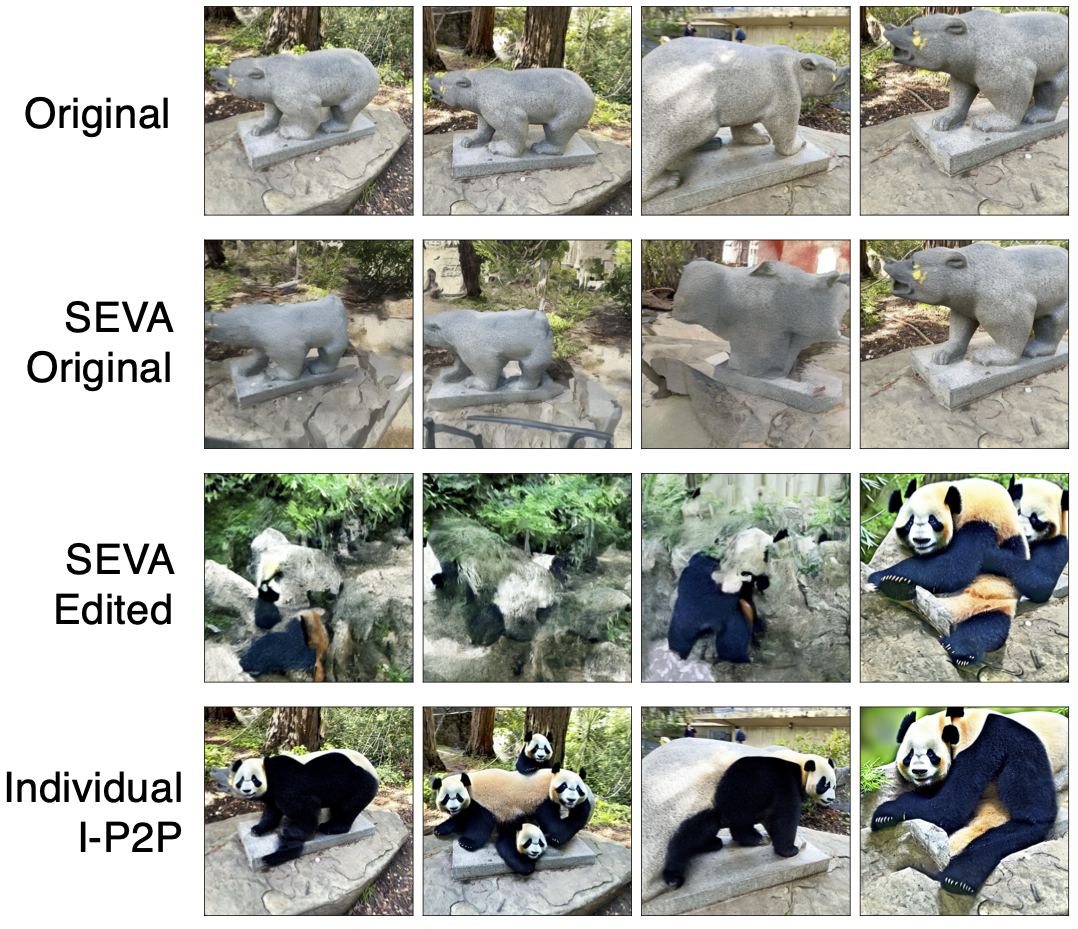}
    \caption{Student and Teacher models limitation example, on the \emph{Bear} scene and \emph{Panda} edit.}
    \label{fig:student_teacher_limitations}
\end{minipage}\hfill
\begin{minipage}{0.48\linewidth}
    \centering
    \includegraphics[width=\linewidth]{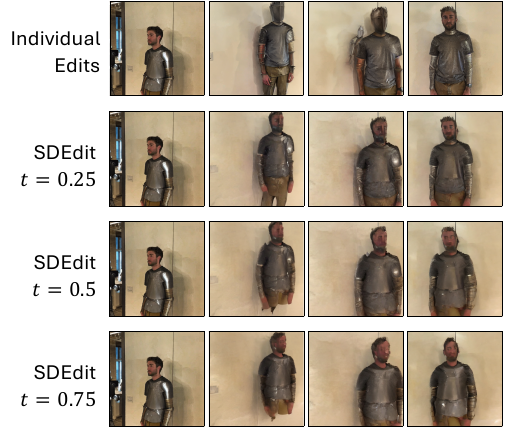}
    \caption{SDEdit failure example, on the \emph{Person} scene and \emph{Knight} edit.}
    \label{fig:sdedit_limitations}
\end{minipage}
\end{figure*}

\section{Extended Qualitative Comparisons with Baselines}
\label{sec:extended_comparison_to_baselines}

In Figures \ref{fig:face_edits}, \ref{fig:face_skull_elf},  \ref{fig:person_edits_1}, \ref{fig:person_edits_2}, \ref{fig:person_edits_3}, 
\ref{fig:person_edits_4},
\ref{fig:bear_edits_1},
\ref{fig:bear_edits_2},
we present additional qualitative comparisons to prior methods, including both enlarged versions of the edits shown in Figure~\ref{fig:comparison_to_baselines} and additional edits. Matching red or purple rectangles highlight regions with multi-view inconsistencies.

\section{Additional Results on Diverse Scenes}
\label{sec:results_for_additional_scenes}

In Figures \ref{fig:car+garden}, \ref{fig:horse+ignatius} we present further qualitative results of \algname\ applied to four different scenes: \emph{Car} from the CO3D dataset~\cite{reizenstein2021common}, \emph{Garden} from the Mip-NeRF 360 dataset~\cite{barron2022mipnerf360}, and \emph{Horse} and \emph{Ignatius} from the Tanks and Temples dataset~\cite{Knapitsch2017}.

\section{Results with More Input Frames}
\label{sec:results_more_frames}

Figure~\ref{fig:8_frames} presents outputs of \algname\ when using $N=8$ input frames.

\section{Beyond Image Editing}
\label{sec:results_beyond_editing}

\algname is not tied to a specific editor or to editing tasks, and can in principle generalize to other multi-view conditional generation scenarios. To illustrate this, we used pre-trained ControlNets \citep{zhang2023adding} as teachers to translate multiple depth or Canny maps of a 3D scene into consistent RGB images. Figure~\ref{fig:canny_and_depth_examples} shows examples. While outputs respect the conditioning and maintain multi-view consistency, they often appear overly blurry, highlighting limitations of SDS-based optimization \citep{poole2022dreamfusion}.

\section{Use of Large Language Models}

Large language models were employed as general-purpose assistants for both writing and coding throughout this work.

\begin{figure*}[!t]
    \centering
    \vspace*{\fill}
    \includegraphics[width=0.6\linewidth]{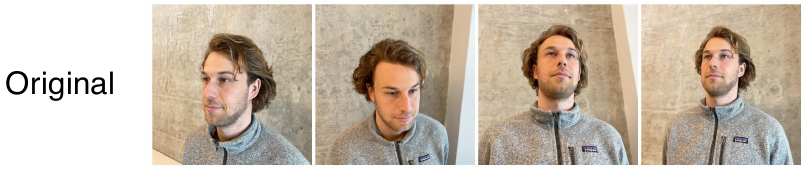}
    \includegraphics[width=0.6\linewidth]{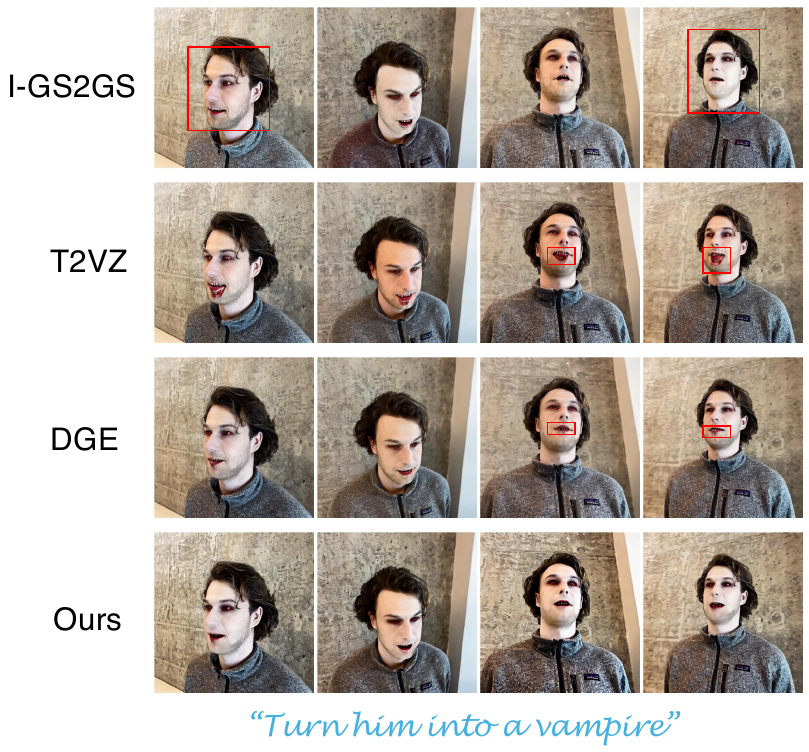}
    \includegraphics[width=0.6\linewidth]{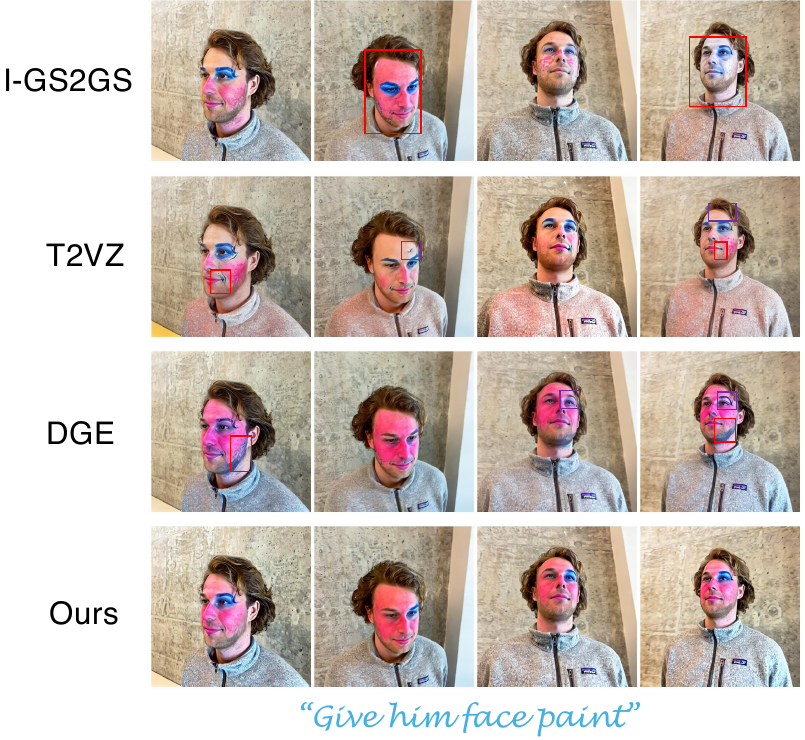}
    \vspace*{\fill}
    \caption{Comparison to baselines on Face scene edits.}
    \label{fig:face_edits}
\end{figure*}
\clearpage

\begin{figure*}[!t]
    \centering
    \vspace*{\fill}
    \includegraphics[width=0.6\linewidth]{figures/face-orig.pdf}
    \includegraphics[width=0.6\linewidth]{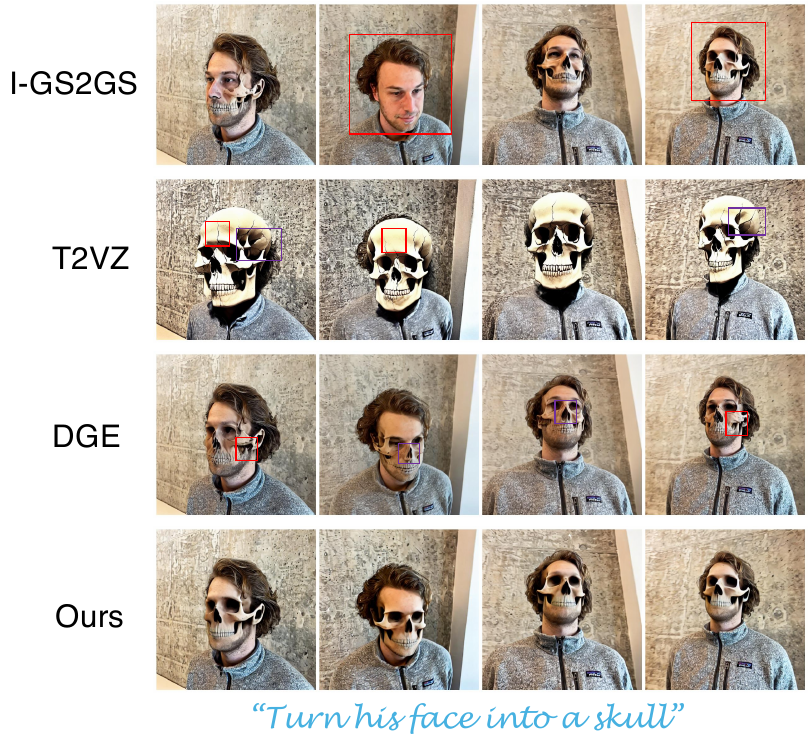}
    \includegraphics[width=0.6\linewidth]{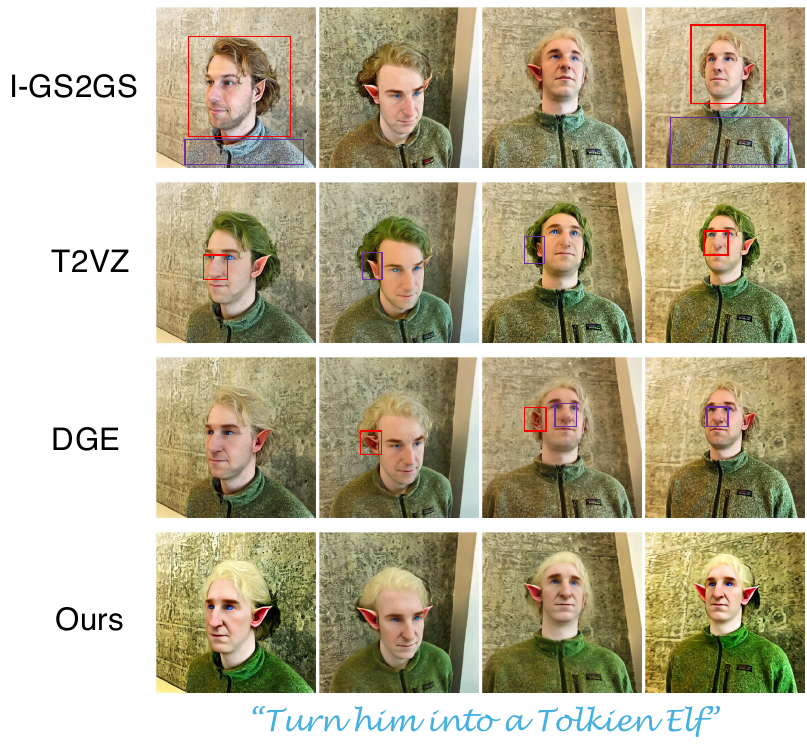}
    \vspace*{\fill}
    \caption{Comparison to baselines on Face scene edits.}
    \label{fig:face_skull_elf}
\end{figure*}
\clearpage

\begin{figure*}[!t]
    \centering
    \vspace*{\fill}
    \includegraphics[width=0.6\linewidth]{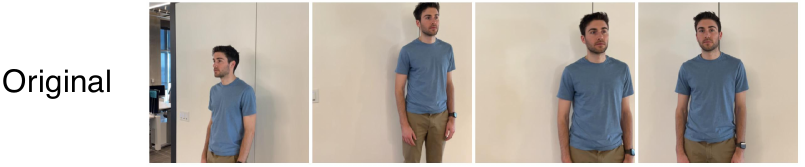}
    \includegraphics[width=0.6\linewidth]{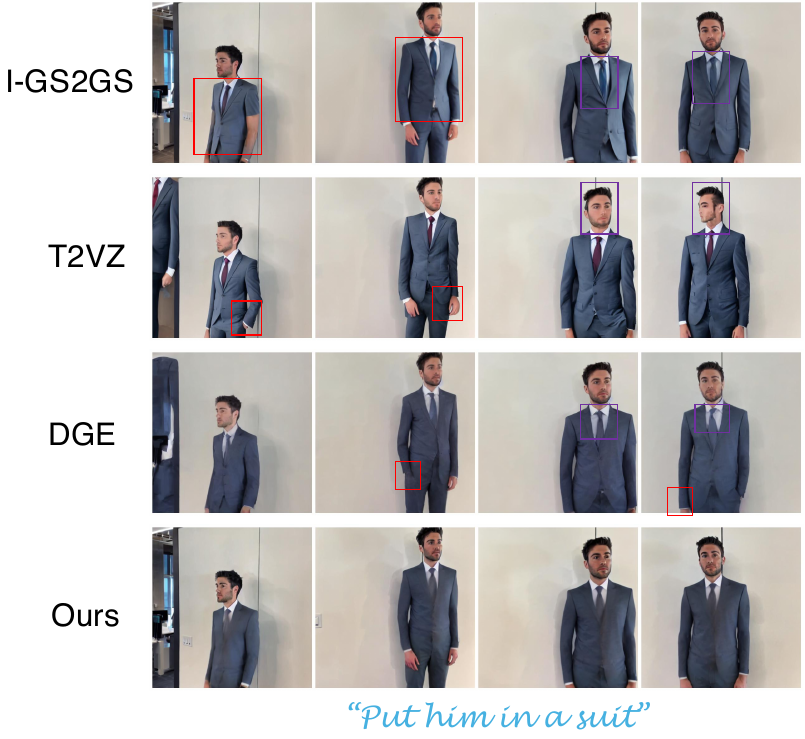}
    \includegraphics[width=0.6\linewidth]{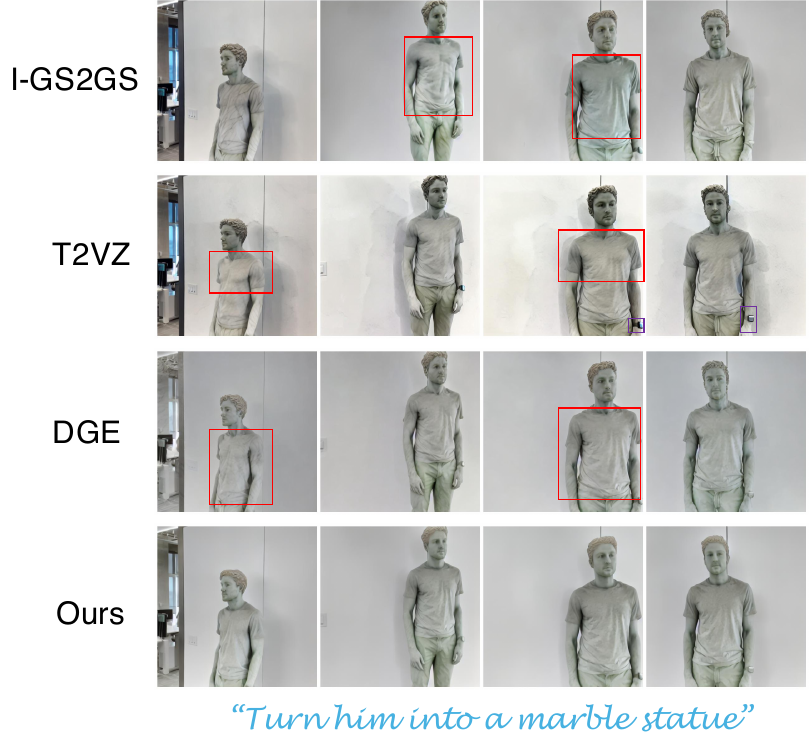}
    \vspace*{\fill}
    \caption{Comparison to baselines on Person scene edits.}
    \label{fig:person_edits_1}
\end{figure*}
\clearpage

\begin{figure*}[!t]
    \centering
    \vspace*{\fill}
    \includegraphics[width=0.6\linewidth]{figures/person-orig.pdf}
    \includegraphics[width=0.6\linewidth]{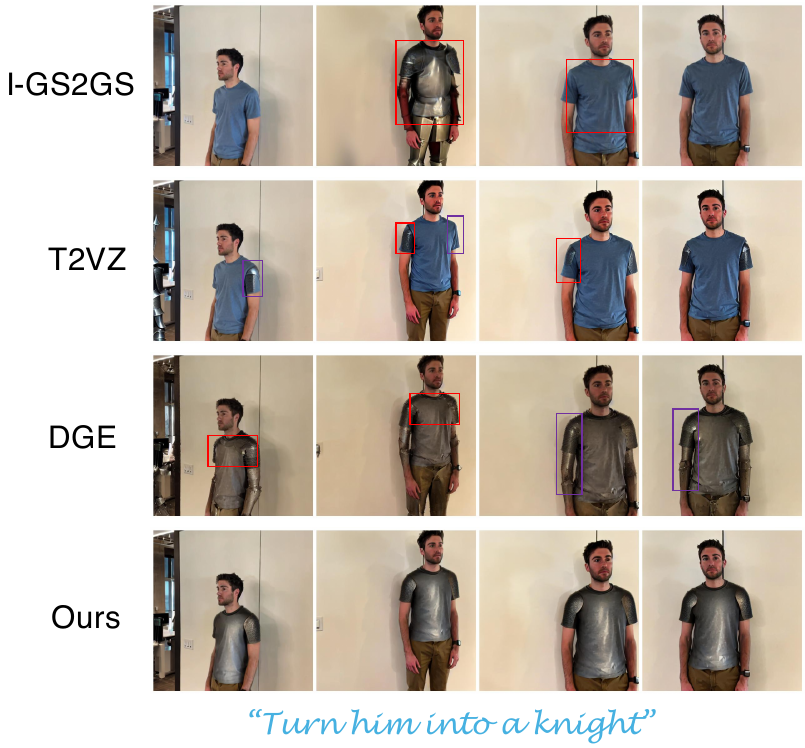}
    \includegraphics[width=0.6\linewidth]{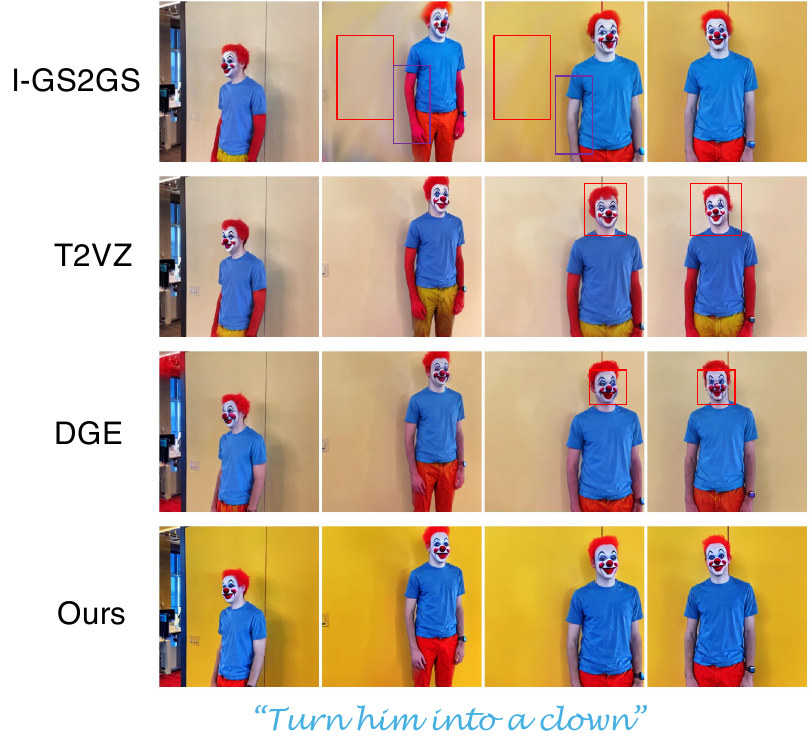}
    \vspace*{\fill}
    \caption{Comparison to baselines on Person scene edits.}
    \label{fig:person_edits_2}
\end{figure*}
\clearpage

\begin{figure*}[!t]
    \centering
    \vspace*{\fill}
    \includegraphics[width=0.6\linewidth]{figures/person-orig.pdf}
    \includegraphics[width=0.6\linewidth]{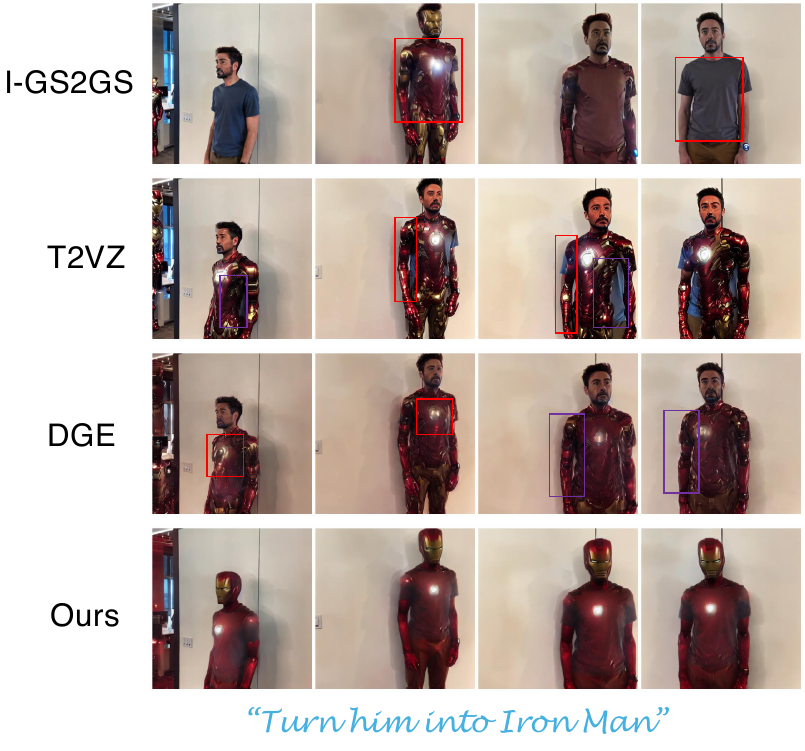}
    \includegraphics[width=0.6\linewidth]{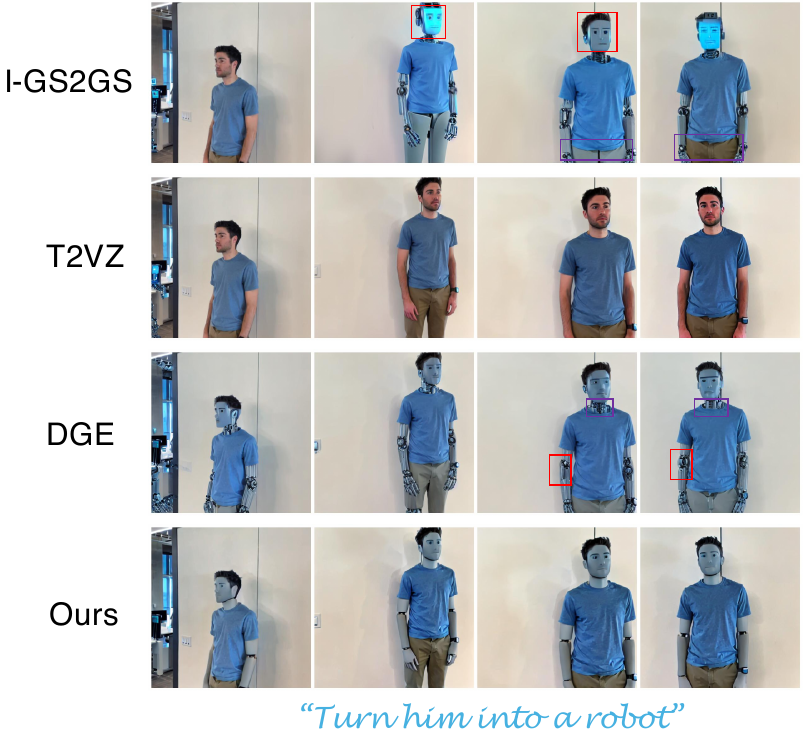}
    \vspace*{\fill}
    \caption{Comparison to baselines on Person scene edits.}
    \label{fig:person_edits_3}
\end{figure*}
\clearpage

\begin{figure*}[!t]
    \centering
    \vspace*{\fill}
    \includegraphics[width=0.6\linewidth]{figures/person-orig.pdf}
    \includegraphics[width=0.6\linewidth]{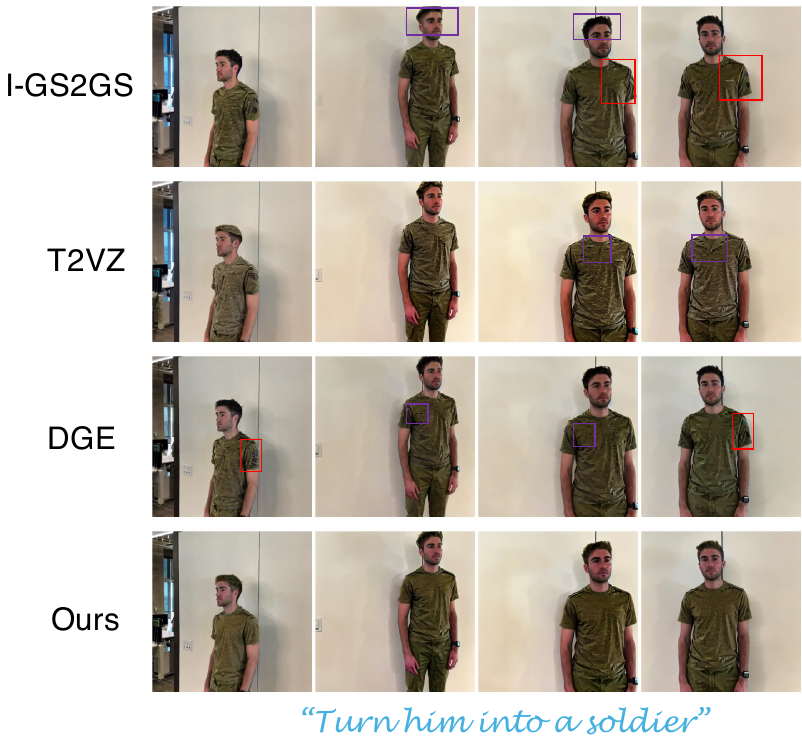}
    \includegraphics[width=0.6\linewidth]{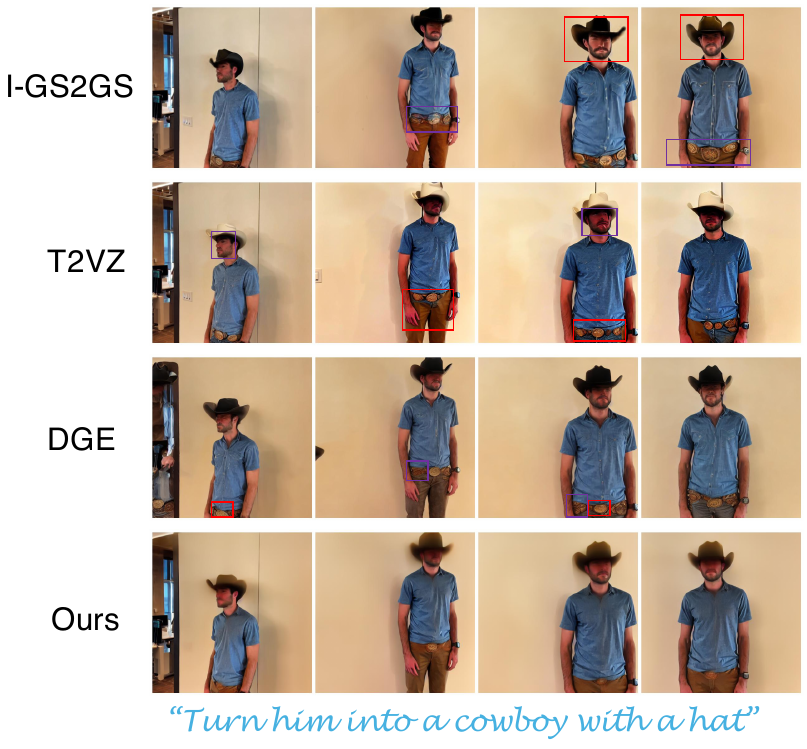}
    \vspace*{\fill}
    \caption{Comparison to baselines on Person scene edits.}
    \label{fig:person_edits_4}
\end{figure*}
\clearpage

\begin{figure*}[!t]
    \centering
    \vspace*{\fill}
    \includegraphics[width=0.6\linewidth]{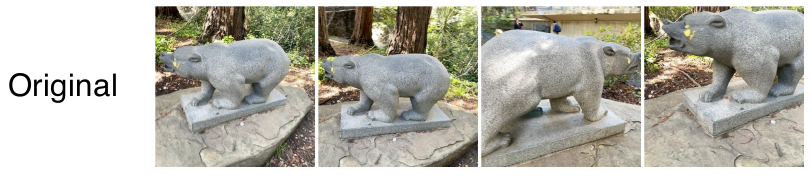}
    \includegraphics[width=0.6\linewidth]{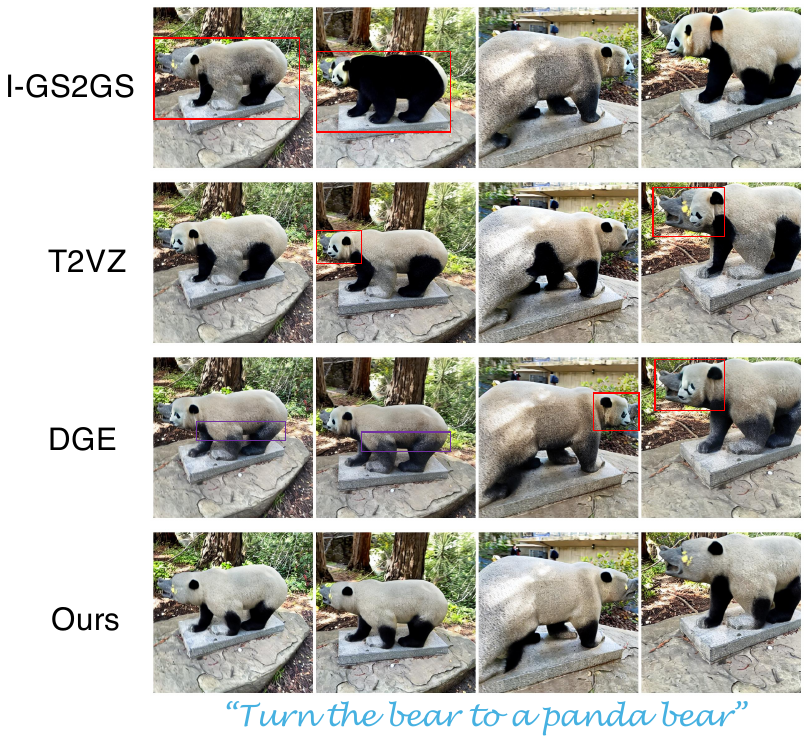}
    \includegraphics[width=0.6\linewidth]{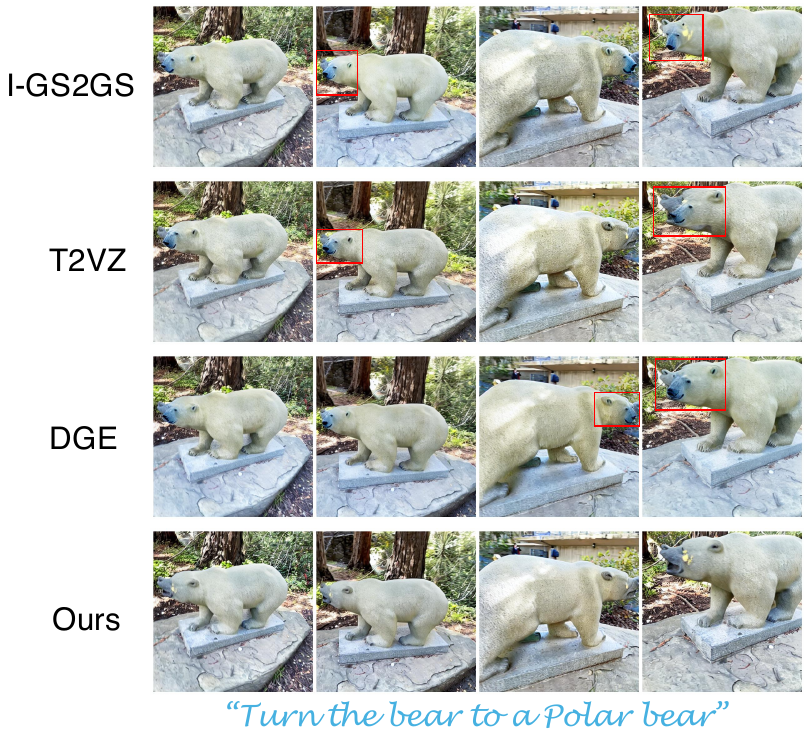}
    \vspace*{\fill}
    \caption{Comparison to baselines on Bear scene edits.}
    \label{fig:bear_edits_1}
\end{figure*}
\clearpage

\begin{figure*}[!t]
    \centering
    \vspace*{\fill}
    \includegraphics[width=0.6\linewidth]{figures/bear-orig.pdf}
    \includegraphics[width=0.6\linewidth]{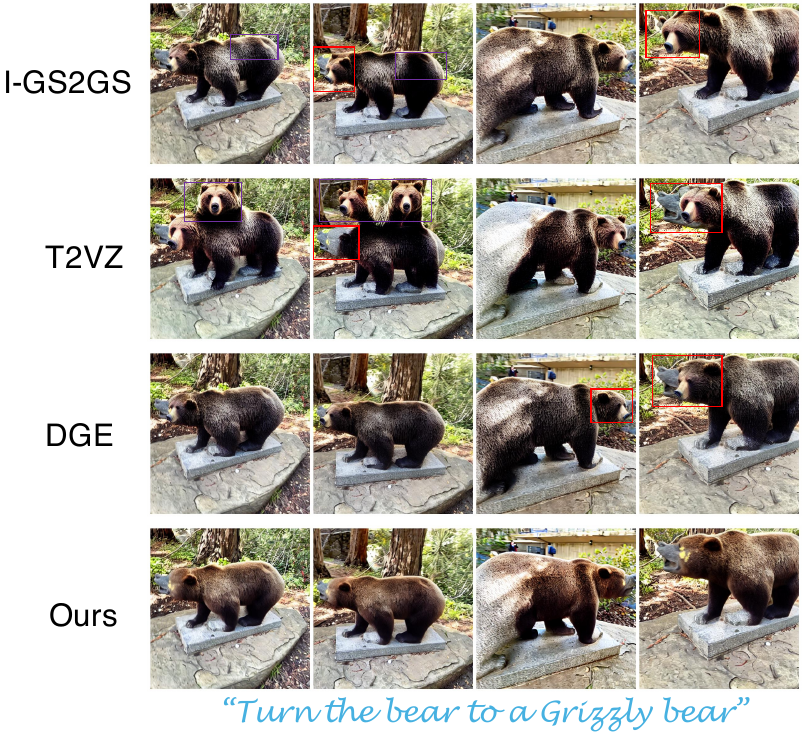}
    \includegraphics[width=0.6\linewidth]{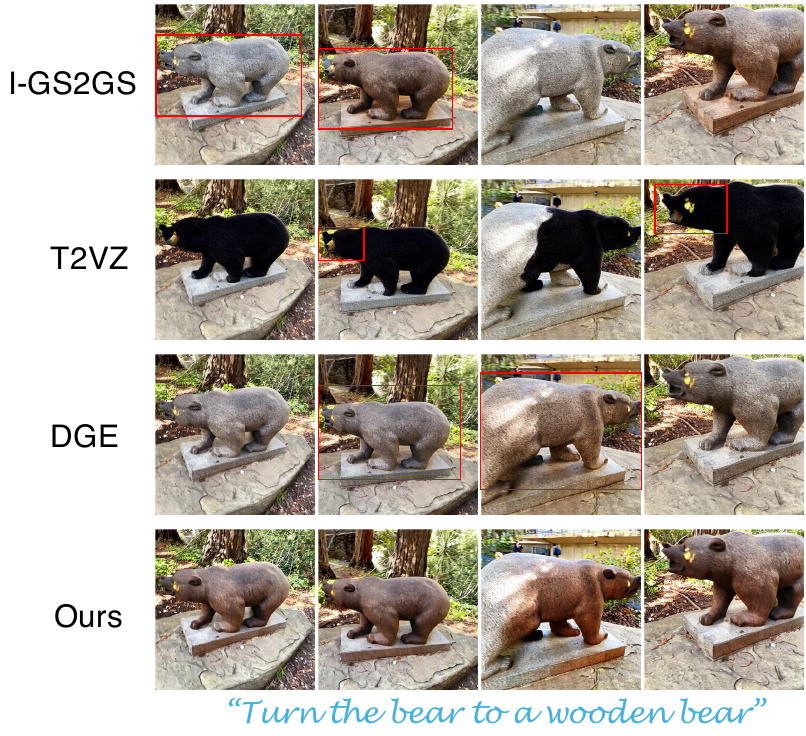}
    \vspace*{\fill}
    \caption{Comparison to baselines on Bear scene edits.}
    \label{fig:bear_edits_2}
\end{figure*}
\clearpage

\begin{figure}
    \centering
    \includegraphics[width=1.0\linewidth]{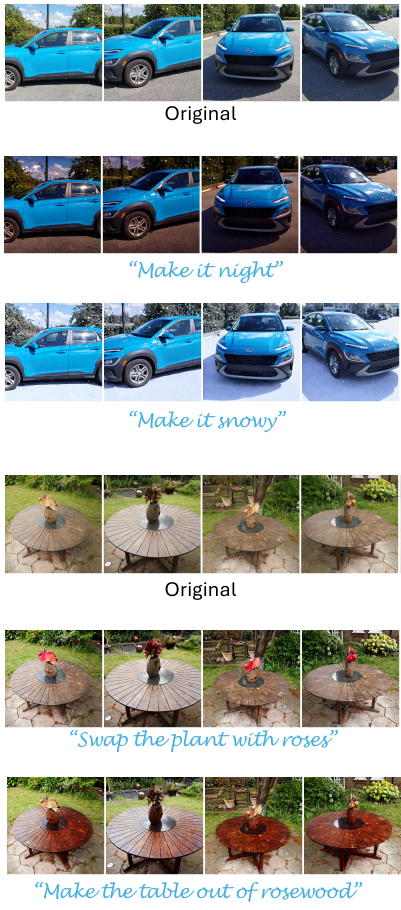}
    \caption{\algname\ edits on the Car (top three rows) and Garden (bottom rows) scenes.}
    \label{fig:car+garden}
\end{figure}

\begin{figure}
    \centering
    \includegraphics[width=1.0\linewidth]{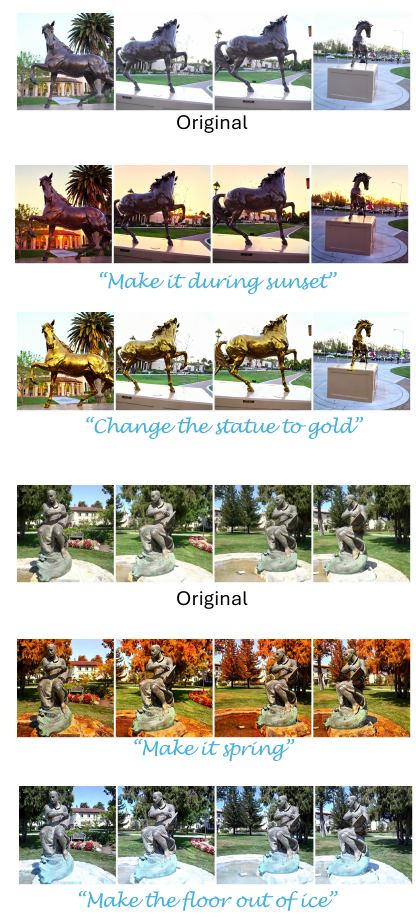}
    \caption{\algname\ edits on the Horse (top three rows) and Ignatius (bottom rows) scenes.}
    \label{fig:horse+ignatius}
\end{figure}

\begin{figure*}
    \centering
    \includegraphics[width=1.0\linewidth]{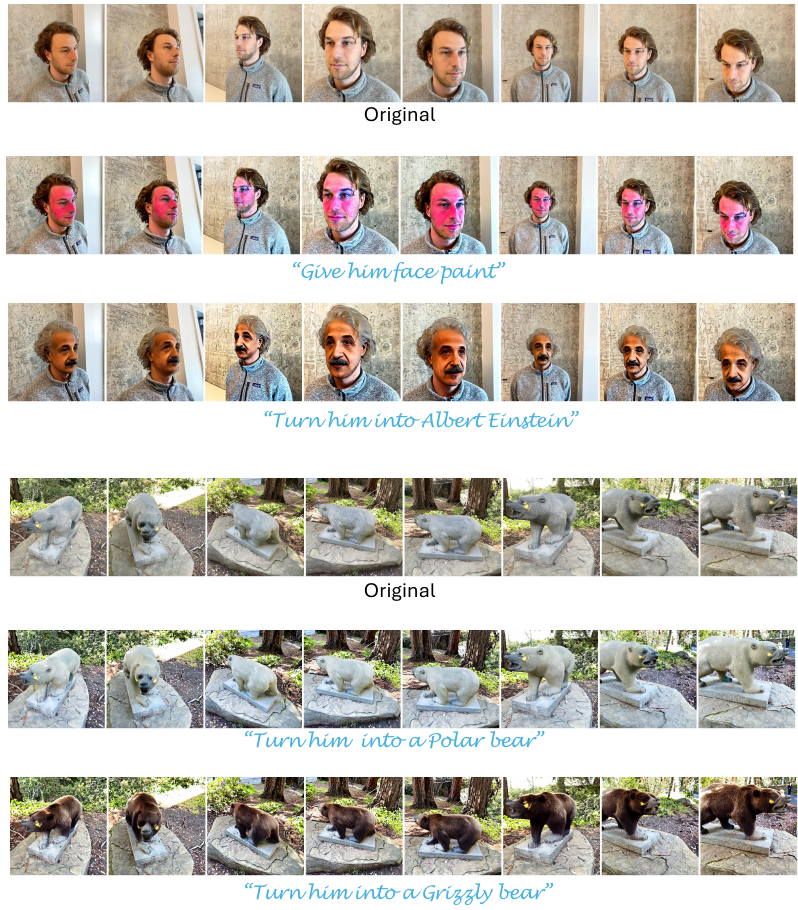}
    \caption{\algname\ edits on 8 input frames on Face and Bear scenes.}
    \label{fig:8_frames}
\end{figure*}

\begin{figure*}
    \centering
    \includegraphics[width=1.0\linewidth]{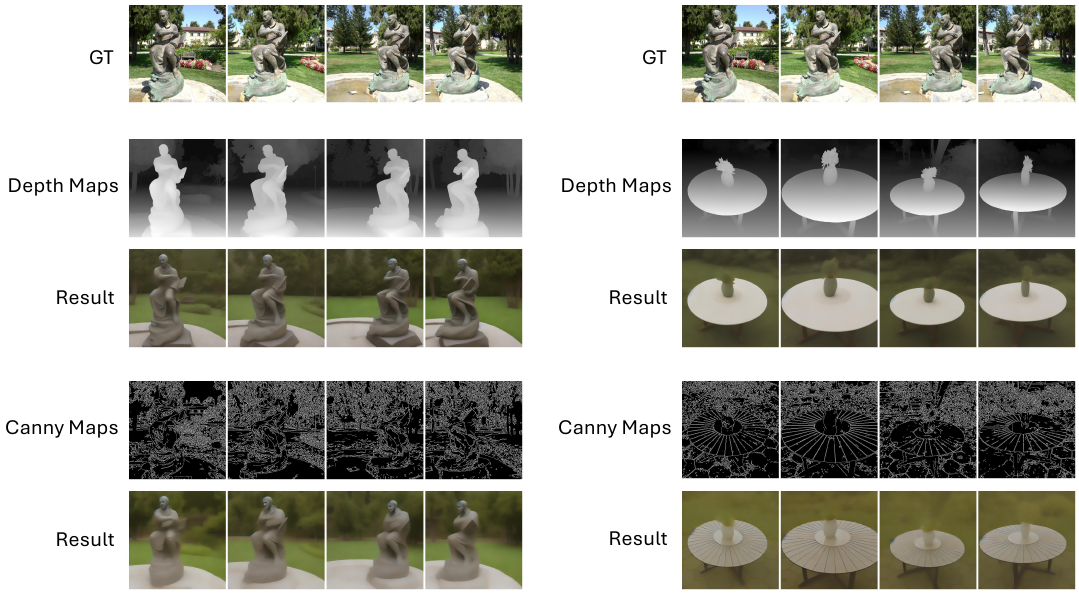}
    \caption{Example results of \algname\ with Canny edge map and Depth maps as input, with corresponding ControlNet teachers.}
    \label{fig:canny_and_depth_examples}
\end{figure*}

\begin{figure*}
\centering
    \includegraphics[width=0.5\linewidth]{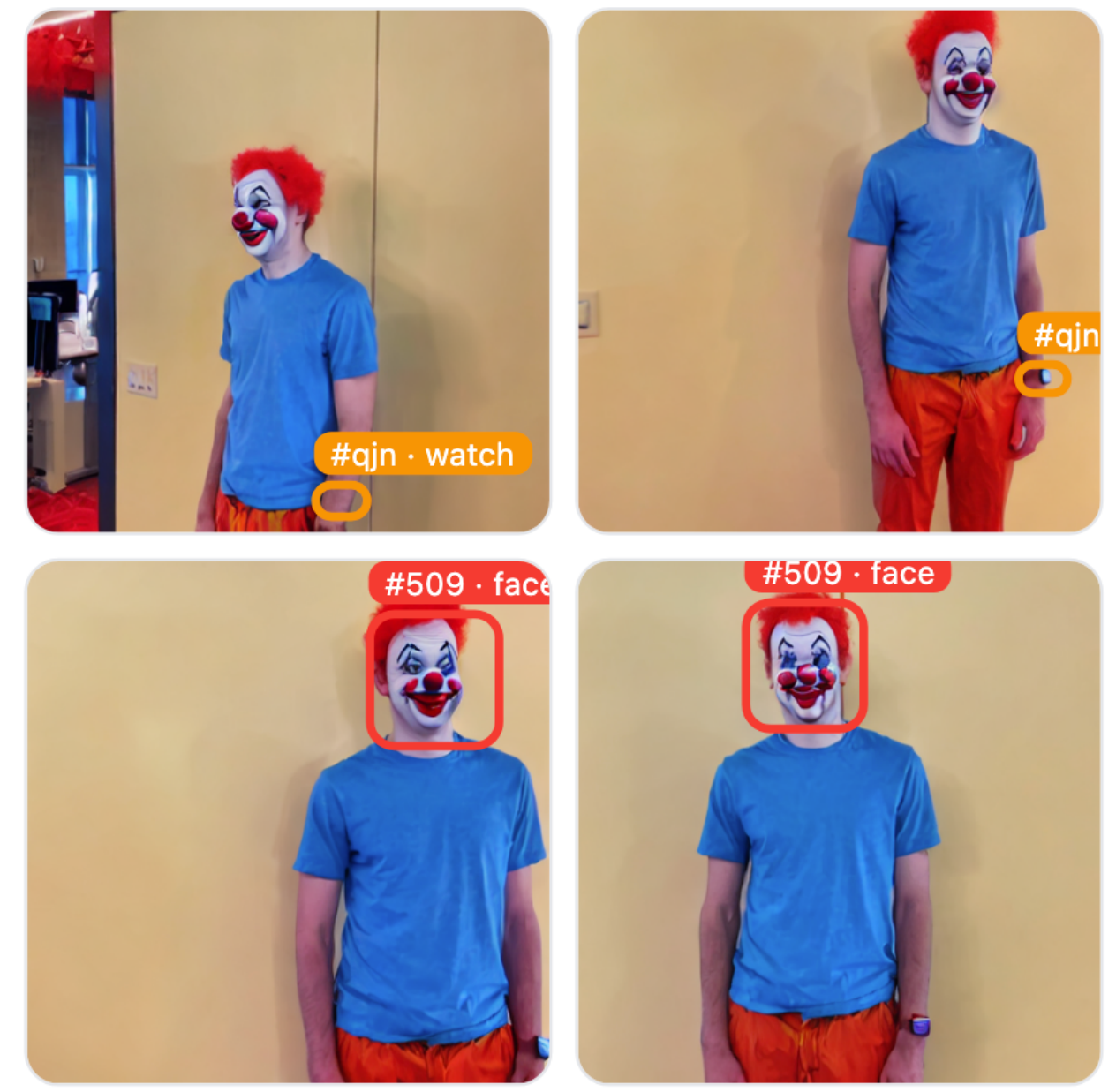}
    \caption{Example of inconsistencies in a scene marked by a human rater.}
    \label{fig:survey_example}
\end{figure*}

\end{document}